\documentclass[twoside]{article}

\usepackage[preprint]{aistats2027}
\usepackage[round]{natbib}

\usepackage{hyperref}
\usepackage{url}
\usepackage{amsmath}
\usepackage{amssymb}
\usepackage{booktabs}
\usepackage{graphicx}
\usepackage{multirow}
\newcommand{\betinn}{\textsc{BeTINN}}

\usepackage{threeparttable}
\usepackage{siunitx}
\usepackage{makecell}
\usepackage{placeins}

\begin{document}

%

\runningtitle{BeTINN for Parametric Stochastic-Volatility Pricing}

%

\twocolumn[

\aistatstitle{Beyond Time Induction: Structured Conditional Neural Solvers for Parametric Stochastic-Volatility Pricing}
\aistatsauthor{
Jose Miguel Lara Rangel$^{1,2}$ \And
Mixtli Vazquez Mendoza$^{2}$
}
\aistatsaddress{
$^{1}$ University College London, UK \\
$^{2}$ Aicraft, Mexico
} ]

\begin{abstract}
Physics-informed neural networks provide a flexible framework for learning parametric PDE solution families without labeled data, yet preserving accurate derivatives across heterogeneous regimes remains challenging. Structured neural solvers offer compact representations, but restrictive conditional parameterizations can limit adaptation across the solution family. We introduce BeTINN, a structured conditional architecture extending time-induced modulation through multi-directional time–parameter conditioning and a joint residual pathway. We evaluate BeTINN on the Heston stochastic-volatility PDE, a demanding parametric problem with mixed derivatives and economically relevant first- and second-order sensitivities. BeTINN improves joint accuracy across normalized option value and key sensitivity metrics relative to our baselines. Controlled ablations characterize the roles of its architectural components. We further find that validation physics loss is informative but does not fully determine differential fidelity, and that accurate solution values can coexist with localized derivative errors. These results support structured conditional adaptation and motivate derivative-aware evaluation when sensitivities are downstream quantities of interest.
\end{abstract}

\section{Introduction}
\label{sec:introduction}

Physics-informed neural solvers provide a flexible framework for approximating differential-equation solutions without requiring large labeled datasets \citep{raissi2017physics,karniadakis2021physics}. Although the literature mainly considers a fixed configuration of physical parameters, many scientific problems require a single surrogate to represent an entire parametric solution family $\mu\mapsto u_\mu$ \citep{penwarden2023metalearning,cho2024parameterized}. In this setting, the network must resolve state and temporal dynamics jointly with variations in the governing parameters, which modify the differential operator across the solution family. The resulting task requires a shared representation that adapts across the geometry of the parametric solution family.

Accurate solution values alone may also be insufficient. Many quantities of scientific and practical interest depend on local differential structure, including sensitivities, fluxes, curvature, and gradients used in optimization or inverse problems \citep{czarnecki2017sobolev,huge2020differential,yu2022gradient}. These quantities can be substantially more sensitive to local approximation error than the solution values themselves:
\[
\begin{aligned}
\|u_{\omega}-u^\star\|\ll 1
\;\not\Rightarrow\;&\;
\|\nabla u_{\omega}-\nabla u^\star\|\ll 1,\\
&\text{or}\quad
\|\nabla^2u_{\omega}-\nabla^2u^\star\|\ll 1 .
\end{aligned}
\]
Consequently, architectures with similar solution-level accuracy can exhibit substantially different differential behavior. For parametric surrogates whose derivatives are quantities of interest, differential fidelity is therefore part of the approximation objective.

A common approach to parametric physics-informed learning concatenates state, time, and physical parameters and processes them jointly with a dense network \citep{penwarden2023metalearning,cho2024parameterized}. Although flexible, this formulation leaves the structure of the parametric solution family to be learned implicitly. More structured representations instead use temporal or parametric variables to adapt the computation performed by the network, promoting parameter sharing across related solutions \citep{de2021hyperpinn,xu2021conditionally}. This introduces a trade-off as overly restrictive adaptation can become a bottleneck, whereas weak conditioning approaches an unconstrained joint-input model and may fail to exploit regularity across the solution family.

We introduce \betinn{}
(\emph{Beyond Time-Induced Neural Network}), a conditional architecture that extends time-induced representations to joint adaptation over time and the governing PDE parameters. \betinn{} uses the governing parameters together with time to modulate the effective network weights. Each modulated parameter group varies along multiple learned adaptation directions, increasing regime-dependent flexibility within a shared structured parameterization. A lightweight joint residual pathway complements this backbone by processing the full problem state directly. The resulting architecture combines structured conditional adaptation with a direct joint representation, while recovering the formulation of \citet{dai2026tinns} as a restricted special case.

We evaluate \betinn{} on stochastic-volatility option pricing under the Heston model, which provides a controlled parametric PDE with demanding differential structure \citep{heston1993closed,rouah2013heston}. A single surrogate must represent variation across the state variables, time to maturity, and a continuous five-dimensional model-parameter space, yielding an eight-dimensional joint input domain. The Heston operator additionally couples first, second, and mixed order derivatives, while several of the corresponding sensitivities are quantities of direct practical interest. Thus, we evaluate normalized option values together with first- and second-order sensitivities, assessing both solution accuracy and local differential structure.

Across our baselines, \betinn{} achieves the strongest overall accuracy for solution and derivative quantities. Ablations identify complementary contributions for the architecture components. Further, we find that solution accuracy and validation physics loss do not fully characterize differential fidelity, motivating explicit derivative-aware evaluation of parametric physics-informed surrogates.

Our contributions are:
\begin{itemize}
    \item We introduce \betinn{}, a conditional architecture for parametric physics-informed learning that jointly adapts network parameters to temporal evolution and governing PDE parameters.

    \item We develop a multi-directional modulation mechanism complemented by a joint residual pathway, increasing regime-dependent flexibility while retaining a shared structured parameterization.

    \item Through controlled comparisons, ablations, and derivative-aware evaluation, we show that \betinn{} improves solution and sensitivity accuracy, while solution and validation-physics metrics do not fully capture differential fidelity.
\end{itemize}

\section{Related Work}
\label{sec:related_work}

\paragraph{Parametric and conditional physics-informed learning.}
PINNs enforce differential equations and boundary conditions through the training objective, with related mesh-free approaches such as DGM \citep{raissi2017physics,sirignano2018dgm}. For parametric PDE families, meta-learning transfers information across instances \citep{penwarden2023metalearning}, P$^2$INNs encode PDE parameters within a shared solver \citep{cho2024parameterized}, and HyperPINN generates solution-network weights from the PDE parameterization \citep{de2021hyperpinn}. Hyper-LR-PINN combines parameter-conditioned meta-learning with shared low-rank weight factors \citep{cho2023hypernetwork}, while conditional networks more broadly adapt model parameters from contextual variables \citep{xu2021conditionally}. TINNs modulate effective network weights as learned functions of time \citep{dai2026tinns}; \betinn{} generalizes this formulation to multi-directional time--parameter adaptation and augments it with a joint residual pathway.

\paragraph{Neural operators and parametric solution maps.}
Neural operators address parametric PDEs by learning mappings between function spaces \citep{kovachki2023neural,lara2026neural}. DeepONet represents nonlinear operators through branch--trunk networks \citep{lu2021learning}, while FNO parameterizes function-space mappings through spectral transformations \citep{li2020fourier}. Physics-informed DeepONets incorporate governing-equation constraints into operator learning and can therefore learn parametric PDE solution operators without paired solution data \citep{wang2021learning}. These methods achieve reuse across PDE instances through operators acting on functional inputs or their representations. Our setting considers a finite-dimensional family of governing parameters and a conditional solution network whose effective computation adapts directly across that parameter space.

\paragraph{Derivative-aware scientific learning.}
Derivative information can also be incorporated explicitly into the training objective. Sobolev training supplements function-value supervision with derivatives of the target function, while differential machine learning similarly exploits derivative information during supervised approximation \citep{czarnecki2017sobolev,huge2020differential}. Gradient-enhanced PINNs additionally penalize derivatives of the PDE residual \citep{yu2022gradient}. We use the same physics-informed supervision across architectures and reserve first- and second-order solution derivatives for evaluation; no target-derivative or residual-gradient terms enter the training objective.

\paragraph{Neural PDE solvers for computational finance.}
Physics-informed surrogates have been used to amortize option-pricing computations across model and contract parameters. For the Heston model, \citet{hainaut2024option} train a single physics-informed surrogate that prices options across multiple parameter settings without retraining. Differential neural surrogates have also been studied for Heston calibration, where accurate sensitivities are directly relevant to the downstream task \citep{zhang2025calibrating}. More recently, DeepSVM uses a physics-informed DeepONet with hard constraints and residual-based adaptive refinement to learn the solution operator across the Heston parameter space \citep{malandain2025deepsvm}. Prior work has also reported noisy derivative-based quantities in the at-the-money regime despite accurate option values, highlighting the gap between solution and derivative accuracy \citep{malandain2025deepsvm}. These studies establish stochastic-volatility pricing as a relevant parametric physics-informed problem.

\begin{figure*}[!t]
    \centering
    \includegraphics[width=0.98\textwidth]{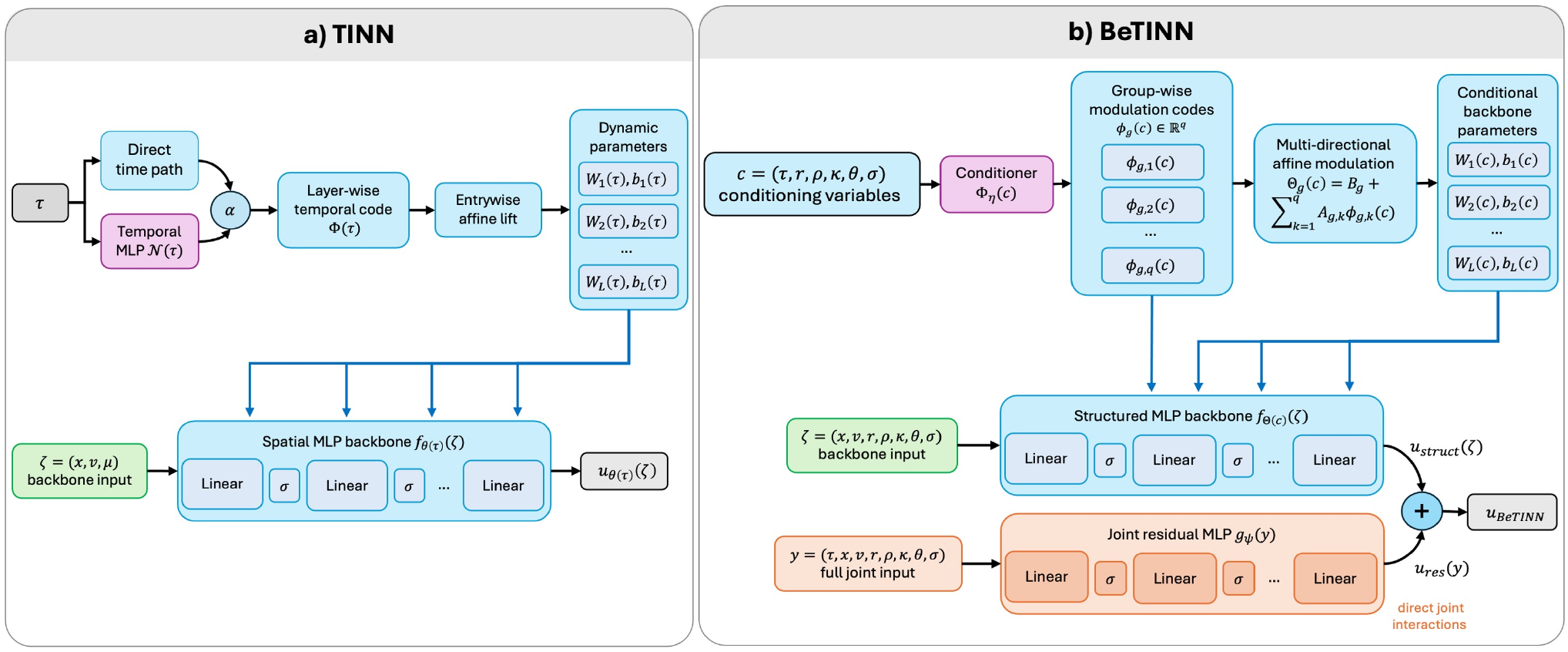}
    \caption{
    \textbf{Architecture Description.}
    \textbf{(a)} TINN uses time-conditioned modulation to adapt backbone parameters over time.
    \textbf{(b)} \betinn{} generalizes time-induced modulation into a structured time--parameter conditional parameterization. Joint conditioning moves each backbone parameter group along multiple learned adaptation directions, yielding a structured family of regime-dependent networks. In parallel, a joint residual pathway captures direct state--time--parameter interactions, and both paths combine to produce the surrogate solution.
    }
    \label{fig:architecture}
    \vspace{-0.8em}
\end{figure*}

\section{Problem Formulation}
\label{sec:problem_formulation}

Our focus is structured conditional learning for parametric time-dependent PDE families. We use the Heston model as a derivative-sensitive application for studying conditional representation design, since its heterogeneous parameter space, mixed differential structure, and economically relevant sensitivities provide a demanding evaluation setting. However, \betinn{} is formulated for general parametric time-dependent PDEs, and Appendix~\ref{app:cross_pde} explores additional parametric PDE families. We therefore introduce the problem at the level of a generic parameter-to-solution map before specializing the formulation to Heston.

\paragraph{Parametric PDE setting.}
We consider a family of time-dependent PDEs indexed by governing parameters $\mu\in\mathcal{P}\subset\mathbb{R}^{p}$, with state $\mathbf{s}\in\Omega\subset\mathbb{R}^{d}$ and temporal coordinate $\tau\in[0,\tau_{\max}]$. For each $\mu$, the solution $u_\mu$ satisfies $\mathcal{F}_{\mu}(u_\mu,\partial_\tau u_\mu, \nabla_{\mathbf{s}}u_\mu,\nabla_{\mathbf{s}}^2u_\mu)=0$ together with the corresponding initial and boundary conditions, defining the parameter-to-solution map $\mathcal{G}:\mathcal{P}\rightarrow\mathcal{U}$, $\mu\mapsto u_\mu$. We seek a single differentiable surrogate $u_\omega(\mathbf{s},\tau,\mu)\approx u_\mu(\mathbf{s},\tau)$ over the joint state--time--parameter domain.

\paragraph{Heston pricing problem.}
We instantiate this setting with the zero-dividend Heston stochastic-volatility model \citep{heston1993closed, rouah2013heston}, where $v$ denotes instantaneous variance and $\mu=(r,\rho,\kappa,\theta,\sigma)\in\mathcal{P}\subset\mathbb{R}^{5}$ contains the risk-free rate, instantaneous correlation, mean-reversion rate, long-run variance, and volatility of variance. For a European call with value $C(S,v,t;\mu)$, strike $K$, and maturity $T$, we define log-moneyness $x=\log(S/K)$, time to maturity $\tau=T-t$, and normalized value $u(\tau,x,v;\mu)=C(S,v,t;\mu)/K$. Under the risk-neutral specification, the normalized pricing equation is
\begingroup
\setlength{\abovedisplayskip}{10pt}
\setlength{\belowdisplayskip}{10pt}
\[
\begin{aligned}
u_\tau=\mathcal{L}_\mu u :=\;&
\tfrac12 v u_{xx}
+\left(r-\tfrac12v\right)u_x \\
&+\rho\sigma v u_{xv}
+\tfrac12\sigma^2v u_{vv} \\
&+\kappa(\theta-v)u_v
-ru .
\end{aligned}
\]
\endgroup
Thus, the surrogate represents the eight-dimensional mapping
$(\tau,x,v,r,\rho,\kappa,\theta,\sigma)\mapsto u(\tau,x,v;\mu)$, coupling
state dependence, temporal evolution, and continuous variation of the
governing dynamics.

\paragraph{Domain and constraints.}
We solve the normalized problem over the window
$\mathcal{D}=[0,\tau_{\max}]\times[x_{\min},x_{\max}]
\times[v_{\min},v_{\max}]\times\mathcal{P}$.
For a European call, we impose the initial condition together with finite-domain
approximations to the asymptotic log-moneyness behavior,
\begingroup
\setlength{\abovedisplayskip}{10pt}
\setlength{\belowdisplayskip}{10pt}
\[
\begin{gathered}
u(0,x,v;\mu)=(e^x-1)^+,
\qquad
u(\tau,x_{\min},v;\mu)=0,\\
u_x(\tau,x_{\max},v;\mu)=e^{x_{\max}}.
\end{gathered}
\]
\endgroup
No explicit boundary penalty is imposed at $v_{\min}$ or $v_{\max}$; the variance direction is constrained by the interior Heston residual over the sampled computational window. Appendix~\ref{app:heston_formulation} discusses this treatment relative to the full-domain Heston problem.

\paragraph{Quantities of interest.}
In addition to the normalized option value, we evaluate differential quantities probing the local structure of the learned solution. Under $S=Ke^x$ and
$C=Ku$,
\begingroup
\setlength{\abovedisplayskip}{10pt}
\setlength{\belowdisplayskip}{10pt}
\[
\begin{aligned}
\Delta
&=\frac{\partial C}{\partial S}
=e^{-x}u_x, \qquad
\frac{1}{K}\frac{\partial C}{\partial v}=u_v \\
K\Gamma
&=K\frac{\partial^2 C}{\partial S^2}
=e^{-2x}(u_{xx}-u_x).
\end{aligned}
\]
\endgroup
These respectively measure first-order price sensitivity, second-order curvature, and sensitivity to the instantaneous variance state. All are obtained by automatic differentiation \citep{baydin2018automatic} of the same surrogate; additional sensitivities are given in Appendix~\ref{app:heston_formulation}.


\section{Beyond Time-Induced Neural Networks}
\label{sec:betinn}

\subsection{Structured Conditional Adaptation}
\label{sec:betinn_structured}

Following the time-induced parameterization of \citet{dai2026tinns}, consider a modulated parameter group $g$ whose effective parameters vary with time as $\Theta_g(\tau)=B_g+A_g\phi_g(\tau)$, where $B_g$ is a learned base tensor, $A_g$ a learned adaptation direction, and $\phi_g(\tau)$ a scalar coefficient from a temporal conditioning network. Thus, $\Theta_g(\tau)-B_g\in\operatorname{span}\{A_g\}$, restricting the time-dependent variation of each group to a single learned direction.

\betinn{} extends this formulation conditioning on $c=(\tau,\mu)$ and allowing multiple adaptation directions:
\begingroup
\setlength{\abovedisplayskip}{8pt}
\setlength{\belowdisplayskip}{8pt}
\[
\Theta_g(c)
=
B_g
+
\sum_{k=1}^{q}
A_{g,k}\phi_{g,k}(c),
\label{eq:betinn_modulation}
\]
\endgroup
where each $A_{g,k}$ has the same shape as $\Theta_g$ and the scalar coefficients $\phi_{g,k}(c)$ are produced jointly by a conditioning network $\Phi_\eta(c)=\{\phi_{g,k}(c)\}_{g,k}$. Conditioning therefore selects coefficients over a shared set of learned adaptation directions. Equivalently, the conditional parameters lie in the affine subspace
\[
\Theta_g(c)-B_g
\in
\operatorname{span}\{A_{g,1},\ldots,A_{g,q}\},
\]
whose dimension is at most $q$. Here, $q$ specifies the maximum number of independent adaptation directions and is not a matrix-rank constraint on the effective weight tensors.

The conditioning variables and backbone inputs play distinct roles. For the
Heston problem, with
$\mu=(r,\rho,\kappa,\theta,\sigma)$, the conditioning vector is
$c=(\tau,\mu)=(\tau,r,\rho,\kappa,\theta,\sigma)$, while the structured backbone receives $\zeta=(x,v,\mu)$. The state variables $(x,v)$ therefore affect the solution through the backbone but do not control its parameter modulation. In contrast, the governing parameters $\mu$ enter both pathways: they directly condition the solution representation and simultaneously adapt the effective backbone parameters. The structured component can therefore be written as $f_{\Theta(c)}(\zeta)$, separating state-dependent evaluation from regime-dependent parameter adaptation while retaining explicit dependence on the governing PDE parameters.

\paragraph{Full \betinn{} Representation.}
The structured modulation confines effective backbone parameters to the conditional affine subspaces above. \betinn{}
complements this representation with a lightweight joint residual network $g_\psi(y)$, with $y=(\tau,x,v,\mu)$, that processes the full problem input directly. The full representation is
\[
u_{\betinn}(\tau,x,v,\mu)
=
f_{\Theta(\tau,\mu)}(x,v,\mu)
+
g_\psi(\tau,x,v,\mu).
\]
The structured backbone provides shared regime-dependent adaptation, while the joint pathway provides a direct channel for state--time--parameter interactions. TINN \citep{dai2026tinns} is recovered by fixing $\mu$, setting $q=1$, using $c=\tau$, and taking $g_\psi\equiv0$.

\betinn{} changes both the structure and dimensionality of the conditional parameterization. The controlled experiments examine these architectural differences through ablations and capacity controls. Additional architectural details are provided in Appendix~\ref{app:betinn_decomposition}, while the corresponding controlled architecture and capacity studies are in Appendix~\ref{app:controlled_ablations}.

\subsection{Physics-Informed Objective and Heston Instantiation}
\label{sec:physics_objective}

Let $\omega$ denote the collection of trainable parameters and $u_\omega$ the differentiable surrogate defined in Section~\ref{sec:problem_formulation}. Training uses no labeled Heston solutions. The governing PDE and the initial and boundary constraints are enforced at sampled points in the joint state--time--parameter domain. For an interior point $\xi=(\tau,x,v,\mu)$, the PDE residual is $
\mathcal{R}_{\mathrm{PDE}}(\xi;\omega)
=
\partial_\tau u_\omega(\xi)
-
\mathcal{L}_\mu u_\omega(\xi),
$
where $\mathcal{L}_\mu$ is the normalized Heston operator defined in
Section~\ref{sec:problem_formulation}. The remaining constraint residuals are
\begingroup
\setlength{\abovedisplayskip}{10pt}
\setlength{\belowdisplayskip}{10pt}
\[
\begin{aligned}
R_{\mathrm{IC}}
&=
u_\omega(0,x,v;\mu)-(e^x-1)^+,\\
R_L
&=
u_\omega(\tau,x_{\min},v;\mu),\\
R_R
&=
\partial_xu_\omega(\tau,x_{\max},v;\mu)-e^{x_{\max}}.
\end{aligned}
\]
\endgroup
Here, $R_{\mathrm{IC}}$ enforces the payoff at maturity, while $R_L$ and
$R_R$ impose the lower- and upper-log-moneyness conditions. No explicit
variance-boundary residual is included in the controlled benchmark.
For each residual block
$j\in\{\mathrm{PDE},\mathrm{IC},L,R\}$ evaluated on its corresponding
point set $\mathcal{C}_j$, we define 
\begingroup
\setlength{\abovedisplayskip}{8pt}
\setlength{\belowdisplayskip}{8pt}
\[
\mathcal{L}_j(\omega)
=
\frac{1}{|\mathcal{C}_j|}
\sum_{\xi\in\mathcal{C}_j}
|R_j(\xi;\omega)|^2,
\qquad
\mathcal{L}_{\mathrm{phys}}
=
\sum_j\mathcal{L}_j.
\]
\endgroup
Thus, each physical constraint contributes its own mean-squared residual, independently of its collocation-set size. All derivatives in the physics objective are computed by automatic differentiation through the full surrogate, including conditional modulation when the differentiated variable enters the conditioner. No solution values or target derivatives from the external reference solver are used during training.

\begin{table*}[t]
\centering
\caption{
Model comparison on the frozen Heston evaluation panel. Entries report mean $\pm$ std over six seeds; lower is better. Baselines are capacity-matched to \betinn{} within $2.4\%$ in trainable parameter count while preserving their native architectural structure.}
\label{tab:main_final_benchmark}

\small
\setlength{\tabcolsep}{5pt}
\begin{tabular*}{\textwidth}{@{\extracolsep{\fill}}lccccc@{}}
\toprule
Method
& Params
& Price MAE $\downarrow$
& $\Delta$ MAE $\downarrow$
& $K\Gamma$ MAE $\downarrow$
& $u_v$ MAE $\downarrow$ \\
\midrule

TINN
& 3,663
& $0.00910 \pm 0.00062$
& $0.0197 \pm 0.0009$
& $0.158 \pm 0.006$
& $0.00853 \pm 0.00046$ \\

PINN
& 3,631
& $0.00857 \pm 0.00054$
& $0.0188 \pm 0.0005$
& $0.149 \pm 0.003$
& $0.00800 \pm 0.00049$ \\

PI-DeepONet
& 3,511
& $0.00974 \pm 0.00063$
& $0.0232 \pm 0.0007$
& $0.199 \pm 0.016$
& $0.0109 \pm 0.0005$ \\

\textbf{\betinn{}}
& 3,595
& $\mathbf{0.00746 \pm 0.00097}$
& $\mathbf{0.0180 \pm 0.0010}$
& $\mathbf{0.143 \pm 0.005}$
& $\mathbf{0.00712 \pm 0.00089}$ \\

\bottomrule
\end{tabular*}
\end{table*}

\begin{figure*}[t]
    \centering
    \includegraphics[width=0.9\textwidth]{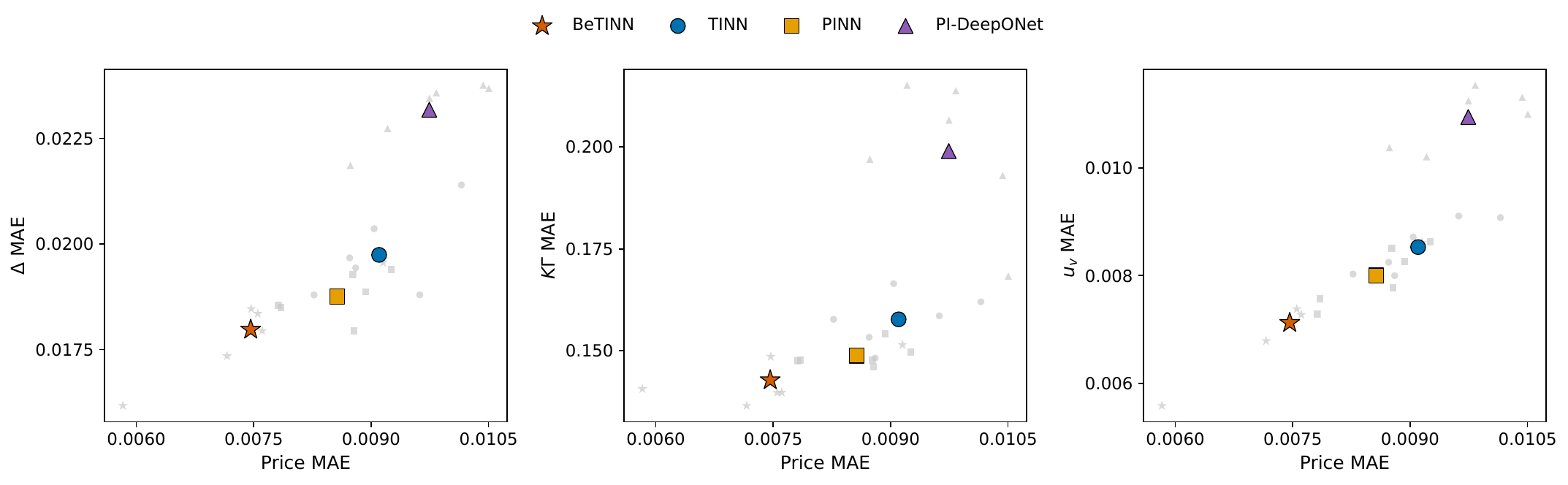}
    \caption{
\textbf{Joint option-value and derivative accuracy.}
Price MAE is plotted against the MAE of $\Delta$, $K\Gamma$, and variance sensitivity $u_v$. Highlighted markers denote the final benchmark models, while light-gray markers show additional controlled configurations. The two sets correspond to their respective frozen evaluation protocols. Lower-left indicates lower error in both option value and the corresponding derivative. \betinn{} consistently occupies the lowest-error region across all three comparisons.
}
    \label{fig:value_derivative_tradeoff}
    \vspace{-0.8em}
\end{figure*}

\section{Experimental Design}
\label{sec:experimental_design}

Controlled models share the physics objective, collocation panels, optimizer, validation criterion, and evaluation sets. Architectural effects are examined through ablations and capacity controls, while method-specific baselines are evaluated separately.

\paragraph{Domain and evaluation panels.} All experiments use the normalized Heston problem from
Section~\ref{sec:problem_formulation} over
\[
\begin{aligned}
\tau&\in[0,5],\quad x\in[-2,2],\quad v\in[0.1,1],\quad \theta\in[0.01,1],\\
r&\in[0,0.2],\; \rho\in[-0.9,0.1],\; \kappa\in[0.5,3],\; \sigma\in[0.1,1].
\end{aligned}
\]
For each collocation set, all unconstrained coordinates are sampled independently and uniformly over their respective ranges, while the coordinate associated with an initial or boundary condition is fixed accordingly.
Interior PDE points use $\tau\in[10^{-6},5]$, whereas the initial-condition panel is sampled separately at $\tau=0$.

Each controlled run uses $15{,}000$ interior points and $100{,}000$ points per initial or boundary constraint, with independent training and validation panels shared across architectures within each seed. Predictive evaluation uses two frozen $10{,}000$-point Heston panels, one for development and one for reported results. Reference values and sensitivities come from a differentiable semi-analytic Heston oracle used only for evaluation. Further sampling details are given in Appendix~\ref{app:sampling_panels}.

\paragraph{Baselines and capacity control.}
We compare \betinn{} with TINN, standard joint-input physics-informed models, and a physics-informed DeepONet baseline. TINN retains temporal-only modulation, with $c=\tau$ and $q=1$, while the governing parameters $\mu$ enter the backbone as standard inputs. Capacity matching controls trainable parameter count; computational cost is evaluated separately. DeepSVM is evaluated separately under its method-specific training procedure in Appendix~\ref{app:deepsvm}. Exact final architectures and parameter counts are reported in Appendix~\ref{app:model_specifications}, the controlled architectural and capacity studies in Appendix~\ref{app:controlled_ablations}, and additional baseline analyses in Appendix~\ref{app:additional_baselines}.

\paragraph{\betinn{} Configuration and Optimization.}
\label{sec:experimental_optimization}
We use $q=4$ adaptation directions, parameter-conditioned modulation, and the joint residual pathway. All controlled architectures are optimized in 64-bit precision with the same Levenberg--Marquardt implementation for at most $3{,}000$ steps \citep{more2006levenberg}. Validation physics loss is evaluated every step with patience $250$, and the best validation checkpoint is restored for evaluation. Model configurations, parameter counts, optimization and checkpoint selection details are in Appendix~\ref{app:model_specifications} and ~\ref{app:optimization_details}.

\paragraph{Evaluation protocol.} We evaluate MAE for the normalized solution $u$, $\Delta$, strike-scaled Gamma $K\Gamma$, and variance sensitivity $u_v$. Metrics are computed over each frozen evaluation panel per run and then aggregated across independent seeds, reporting mean and standard deviation. Extended global and regional results are in Appendix~\ref{app:final_benchmark}, local and spatial differential diagnostics in ~\ref{app:differential_diagnostics}, and shape and sign constraint diagnostics in ~\ref{app:shape_diagnostics}. Metric and diagnostic definitions are provided in Appendix~\ref{app:extended_evaluation_metrics}.

\begin{figure*}[t]
    \centering
    \includegraphics[width=0.95\textwidth]{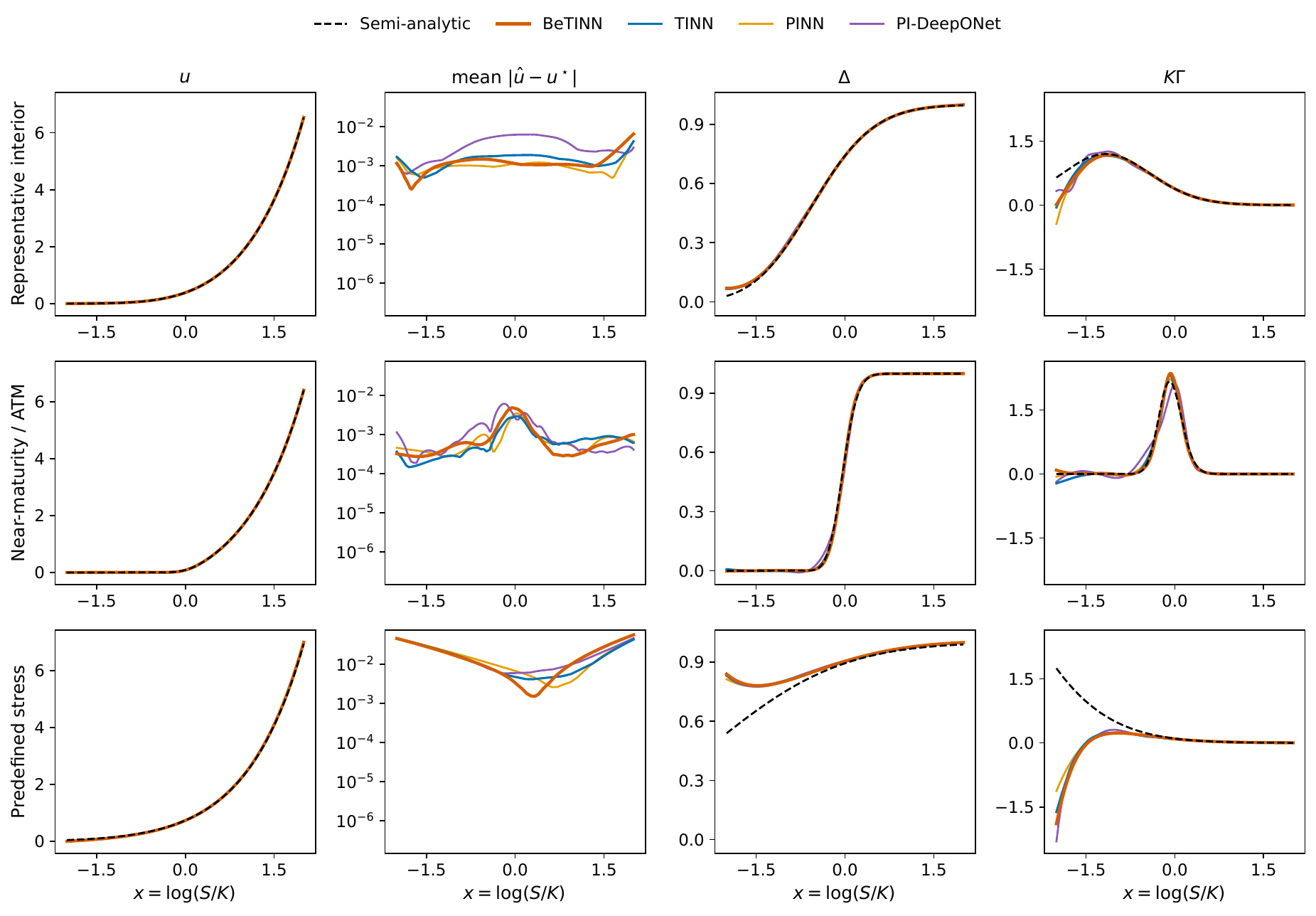}
    \caption{
\textbf{Solution and derivative profiles across representative Heston regimes.}
We sweep log-moneyness $x$ over three fixed $(\tau,v,r,\rho,\kappa,\theta,\sigma)$ settings: interior $(1.84,0.52,0.08,-0.38,1.92,0.39,0.55)$, near-maturity $(0.07,0.57,0.16,-0.48,1.61,0.40,0.75)$, and stress $(3.78,0.64,0.15,-0.47,1.89,0.99,0.60)$. While all models closely match the price profiles, sensitivity errors are more visible in high-curvature and low-moneyness regions. The near-maturity slice shows differences around the sharp $K\Gamma$ peak. Across the frozen evaluation panel, low-moneyness growth in $\Delta$ and $K\Gamma$ errors is amplified by the coordinate factors $e^{-x}$ and $e^{-2x}$; in the long-maturity stress, finite-domain boundary effects may also contribute near the computational boundary.}
\label{fig:solution_derivative_profiles}
\vspace{-0.8em}
\end{figure*}

\begin{figure*}[t]
    \centering
    \includegraphics[width=0.98\textwidth]{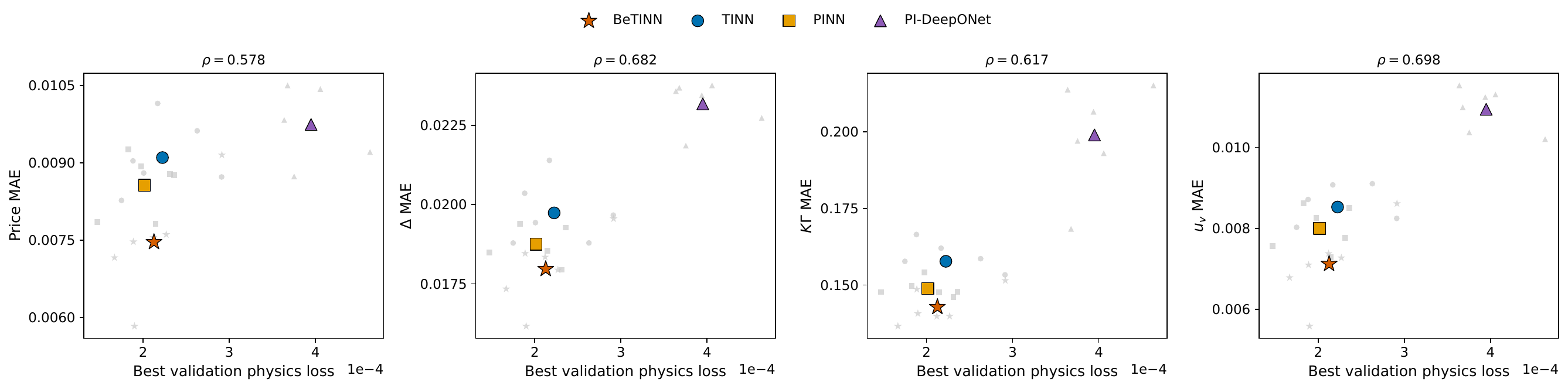}
    \caption{
\textbf{Validation physics loss and downstream predictive error.}
Best validation physics loss is compared with MAE for price, $\Delta$, $K\Gamma$, and variance sensitivity $u_v$. Highlighted markers denote the 24 final benchmark runs from 4 architectures and 6 confirmatory seeds. Spearman correlations, reported above each panel, are computed over these model--seed pairs. Light-gray markers show additional controlled variants for visual context and are excluded from the correlation calculation. Validation physics loss is positively associated with downstream error, but does not preserve the same ranking across solution and derivative quantities.
}
    \label{fig:validation_downstream_proxy}
    \vspace{-0.9em}
\end{figure*}

\section{Results}
\label{sec:results}

We evaluate whether the proposed conditional parameterization improves solution accuracy and the differential quantities derived from the learned surrogate. The main results examine joint value--derivative accuracy under controlled capacity, the relationship between validation physics loss and downstream fidelity, and the localization of the remaining derivative errors.

\subsection{Joint Solution and Derivative Accuracy}
\label{sec:results_main}

Table~\ref{tab:main_final_benchmark} shows that \betinn{} achieves the lowest MAE across the solution, first-order sensitivities, and second-order curvature. Improvements vary across observables, with larger gains in price and $u_v$ and more modest gains in $\Delta$ and $K\Gamma$. The comparison uses closely matched parameter budgets, strengthening attribution of these gains to the conditional representation. Their consistency across observables of different differential order indicates more accurate approximation of the local differential quantities evaluated here. Paired seed comparisons show the same overall pattern across the principal baselines (Appendix~\ref{app:final_seed_robustness}). Controlled development ablations show that none of the tested partial variants reproduces the joint accuracy of the full \betinn{} configuration (Appendix~\ref{app:controlled_ablations}).

Figure~\ref{fig:value_derivative_tradeoff} complements the aggregate benchmark by examining value and derivative errors jointly. \betinn{} lies in the low-error region in each price--derivative plane, showing that its lower option-value error is accompanied by low error in $\Delta$, $K\Gamma$, and $u_v$. The comparison does not indicate an evident trade-off between improving the solution value and preserving its differential structure. Further, the controlled development variants also show that similar price accuracy can correspond to different derivative errors, motivating direct evaluation of the sensitivities.

\subsection{Validation Physics Loss and Downstream Fidelity}
\label{sec:results_validation}
Model selection in our experiments uses only held-out physics loss, without access to the semi-analytic reference. Figure~\ref{fig:validation_downstream_proxy} examines how well this criterion reflects the downstream quantities used for evaluation. Validation physics loss is positively associated with all four predictive errors, with Spearman correlations ranging from 0.578 to 0.698 across the final benchmark model–seed pairs. These moderate associations indicate that held-out physics loss is informative of downstream accuracy, although architectures with similar validation loss can differ in their solution and derivative errors.

\subsection{Derivative Errors Concentrate in Challenging Local Regimes}
\label{sec:results_profiles}

Global MAE does not reveal where differential errors arise. Figure~\ref{fig:solution_derivative_profiles} examines fixed Heston slices across interior, near-maturity, and stress regimes. In all three, option-value profiles remain close to the semi-analytic solution, while derivative errors are more pronounced. Near maturity, discrepancies concentrate around the sharp $K\Gamma$ peak, where curvature is strongest. Across the frozen evaluation panel, low-moneyness $\Delta$ and $K\Gamma$ errors are substantially amplified by the factors $e^{-x}$ and $e^{-2x}$ transforming $u_x$ and $u_{xx}-u_x$ into financial sensitivities. In the long-maturity stress regime, finite-domain boundary approximations may also contribute near the boundary. Overall, the local diagnostics show that small solution error does not necessarily imply equally accurate differential quantities.

Appendix~\ref{app:regional_results} explores how regional performance varies across observables and local regimes. Appendix~\ref{app:differential_diagnostics} extends this analysis with pointwise derivative-error profiles, two-dimensional $(x,v)$ error maps, and spatial PDE-residual diagnostics.

\section{Discussion}
\label{sec:discussion}

\paragraph{Structured conditional adaptation.}
The benchmark and controlled ablations consistently support the importance of how conditional adaptation is organized in \betinn{}. Multi-directional modulation, parameter conditioning, and the joint pathway contribute jointly, indicating that performance depends on how conditional capacity is organized across the solution family.

\paragraph{Physics consistency and differential fidelity.}
The results distinguish physics consistency from differential fidelity. Validation physics loss remains informative of downstream accuracy, although its association varies across observables and does not exactly preserve model rankings. The Heston residual couples multiple derivatives, whereas $\Delta$, $K\Gamma$, and $u_v$ probe specific aspects of the local solution geometry. Differential accuracy is therefore important when downstream use depends on sensitivities, as in Heston applications such as hedging and risk analysis. More generally, the same consideration applies to derivative-dependent tasks including sensitivity analysis, inverse problems, optimization, and control.

\paragraph{Scope and limitations.}
Our primary analysis focuses on the parametric Heston family. Although \betinn{} is not specific to option pricing and Appendix~\ref{app:cross_pde} evaluates additional parametric PDE families, broader validation across operators, dimensionalities, and regularity regimes remains necessary. The semi-analytic Heston solution also enables accurate evaluation of sensitivities, which may be harder to obtain in other PDE settings.

Capacity matching controls model size while leaving differences in function class, optimization geometry, and computational cost. The benchmark supports improved solution and derivative accuracy under closely matched parameter budgets. Derivative fidelity is assessed empirically, without formal guarantees of monotonicity, convexity, or arbitrage consistency beyond the physics constraints. Shape and sign diagnostics are reported in Appendix~\ref{app:shape_diagnostics}.

\paragraph{Future work.}
Open directions include adaptive conditional subspaces across parameter space, derivative-aware physics-informed objectives without labeled sensitivities, and validation on higher-dimensional and less regular PDE families. These directions are particularly relevant when learned derivatives are themselves important outputs, including sensitivity analysis, inverse problems, optimization, and control.

\section{Conclusion}
\label{sec:conclusion}

We introduced \betinn{}, a structured conditional architecture for parametric physics-informed learning that extends time-induced modulation to joint time--parameter conditioning with multiple adaptation directions and a complementary residual pathway. On the parametric Heston PDE, \betinn{} achieves the lowest mean MAE across the normalized solution, first-order sensitivities, and second-order curvature among the capacity-controlled models considered. Ablation analyses further support the importance of how conditional representation is organized across the solution family. These results support structured conditional adaptation for parametric PDE surrogates with demanding differential structure.

More broadly, physics consistency, solution accuracy, and differential fidelity capture complementary aspects of surrogate quality. Validation physics loss remains informative of downstream quality, while local profile analysis reveals derivative errors that are weakly reflected in solution accuracy. These findings motivate treating derivative fidelity as an essential evaluation criterion when sensitivities are part of the downstream scientific or computational objective.


\clearpage

\bibliography{aistats}

\clearpage
\appendix
\thispagestyle{empty}

\onecolumn
\aistatstitle{BeTINN Supplementary Materials}

\section{Additional Heston Formulation Details}
\label{app:heston_formulation}

\paragraph{Risk-neutral dynamics.}
Under the risk-neutral measure and assuming zero dividends, the Heston model \citep{heston1993closed, rouah2013heston} specifies the asset price $S_t$ and instantaneous variance $v_t$ as
\[
\begin{aligned}
dS_t
&=
rS_t\,dt
+
S_t\sqrt{v_t}\,dW_t^{S},\\
dv_t
&=
\kappa(\theta-v_t)\,dt
+
\sigma\sqrt{v_t}\,dW_t^{v},
\end{aligned}
\qquad
d\langle W^{S},W^{v}\rangle_t=\rho\,dt .
\]
The governing parameters are $\mu=(r,\rho,\kappa,\theta,\sigma)$, where $r$ is the risk-free rate, $\rho\in[-1,1]$ the instantaneous correlation, $\kappa>0$ the mean-reversion rate, $\theta>0$ the long-run variance, and $\sigma>0$ the volatility of variance.

For a European contingent claim with value $C(S,v,t;\mu)$, absence of arbitrage yields the backward pricing equation
\[
\partial_t C
+
\frac{1}{2}vS^2\partial_{SS}C
+
\rho\sigma vS\,\partial_{Sv}C
+
\frac{1}{2}\sigma^2v\,\partial_{vv}C
+
rS\,\partial_SC
+
\kappa(\theta-v)\partial_vC
-
rC
=
0.
\]

\paragraph{Normalized coordinates.}
With
\[
x=\log(S/K),
\qquad
\tau=T-t,
\qquad
u(\tau,x,v;\mu)=\frac{C(S,v,t;\mu)}{K},
\]
the relevant derivatives satisfy
\[
\begin{aligned}
\partial_S C
&=e^{-x}u_x,
&
\partial_{SS}C
&=\frac{e^{-2x}}{K}(u_{xx}-u_x),
&
\partial_{Sv}C
&=e^{-x}u_{xv},\\
\partial_vC
&=Ku_v,
&
\partial_{vv}C
&=Ku_{vv},
&
\partial_tC
&=-Ku_\tau .
\end{aligned}
\]
Substitution into the backward equation yields the normalized Heston PDE reported in Section~\ref{sec:problem_formulation}.

\paragraph{Computational-domain constraints.}
For a European call, the terminal payoff becomes the initial condition
\[
u(0,x,v;\mu)=(e^x-1)^+
\]
in time-to-maturity coordinates. At the finite log-moneyness boundaries, we
use the asymptotic approximations
\[
u(\tau,x_{\min},v;\mu)=0,
\qquad
u_x(\tau,x_{\max},v;\mu)=e^{x_{\max}}.
\]

An oracle audit confirms that these approximations are highly accurate at short
maturities but become less accurate for long-maturity, high-variance regimes. At $x_{\min}=-2$, the mean full-domain oracle mismatch is $9.9\times10^{-3}$, increasing to $4.4\times10^{-2}$ for $\tau\ge3.75$ and $\theta\ge0.7525$. At $x_{\max}=2$, the mean discrepancy from $\Delta=1$ is $5.0\times10^{-3}$ and increases to $1.2\times10^{-2}$ in the same regime. These finite-domain effects are therefore relevant when interpreting absolute errors near the log-moneyness
boundaries.

No explicit boundary penalty is imposed at either $v_{\min}$ or $v_{\max}$ in the controlled experiments. The finite variance interval is therefore used as a computational sampling window, with the Heston differential operator enforced through the interior PDE residual over the sampled range. This treatment is motivated by related physics-informed Heston formulations \citep{malandain2025deepsvm}, although our lower truncation at $v_{\min}=0.1$ is nondegenerate and does not inherit the boundary classification of the original Heston boundary at $v=0$. Accordingly, we do not interpret the finite variance interval as a classically closed truncated boundary-value problem. Performance near both variance edges is evaluated explicitly in Appendix~\ref{app:regional_results}.

\paragraph{Additional sensitivities.}
Beyond $\Delta$, strike-scaled Gamma $K\Gamma$, and variance sensitivity $u_v$ reported in the primary benchmark, the extended evaluation includes sensitivity with respect to instantaneous spot volatility,
\[
\frac{1}{K}\frac{\partial C}{\partial\sqrt{v}}
=
2\sqrt{v}\,u_v,
\]
normalized calendar Theta,
\[
\frac{1}{K}\frac{\partial C}{\partial t}
=
-u_\tau,
\]
and normalized interest-rate sensitivity,
\[
\frac{1}{K}\frac{\partial C}{\partial r}
=
u_r.
\]
All sensitivities are computed by automatic differentiation of the same trained surrogate; no separately trained derivative models or finite-difference approximations are used.

\section{Additional \betinn{} Architecture Details}
\label{app:betinn_decomposition}

\paragraph{Conditioning-network output.}
For $G$ independently modulated parameter groups and $q$ adaptation directions per group, the conditioning network produces $Gq$ scalar coefficients,
\[
\Phi_\eta(c)
=
\left\{
\phi_{g,k}(c)
\right\}_{g=1,\ldots,G;\,k=1,\ldots,q}.
\]
For each parameter group $g$, these coefficients determine the conditional
combination of a shared set of learned adaptation tensors,
\[
\Theta_g(c)
=
B_g
+
\sum_{k=1}^{q}
A_{g,k}\phi_{g,k}(c).
\]
The tensors $B_g$ and $A_{g,k}$ are shared across all conditioning configurations, while only the coefficients $\phi_{g,k}(c)$ vary with $c$.
Thus, the conditioner does not generate an unrestricted set of backbone parameters; instead, it selects coordinates within a learned conditional parameterization.

\paragraph{Architectural decomposition.}
The \betinn{} architecture separates three independently controllable mechanisms. First, the number of adaptation directions $q$ controls the dimensionality of the conditional modulation space for each parameter group. Second, the conditioning variables determine whether the effective parameters depend only on time, $c=\tau$, or jointly on time and the governing parameters, $c=(\tau,\mu)$. Third, the joint residual pathway controls whether the full state--time--parameter input is processed directly through $g_\psi(\tau,x,v,\mu)$ in addition to the structured backbone.

The TINN construction \citep{dai2026tinns} is recovered as a restricted case by fixing the governing parameters $\mu$, setting $q=1$, using temporal-only conditioning $c=\tau$, and removing the joint residual pathway, $g_\psi\equiv0$. The controlled ablations in the following sections vary these mechanisms individually and in combination to examine their contributions to the learned representation.

\section{Experimental Configuration and Reproducibility}
\label{app:experimental_configuration}

\subsection{Sampling and Fixed Panels}
\label{app:sampling_panels}

All experiments use the normalized Heston problem over
\[
\begin{aligned}
\tau&\in[0,5], &
x&\in[-2,2], &
v&\in[0.1,1], &
r&\in[0,0.2],\\
\rho&\in[-0.9,0.1], &
\kappa&\in[0.5,3], &
\theta&\in[0.01,1], &
\sigma&\in[0.1,1].
\end{aligned}
\]
Interior PDE collocation uses $\tau\in[10^{-6},5]$, whereas initial-condition points are sampled separately at $\tau=0$. Unless otherwise specified, all unconstrained coordinates are sampled independently and uniformly over their corresponding ranges.

The log-moneyness boundary panels fix the coordinate associated with the imposed constraint while sampling the remaining coordinates independently from the same distributions. No separate variance-boundary training panel is used in the controlled experiments. The fixed training panels are summarized in Table~\ref{tab:training_panels}.

\begin{table}[t]
    \centering
    \caption{
    Physics-informed training panels used in the controlled experiments.
    }
    \label{tab:training_panels}
    \begin{tabular}{lcc}
        \toprule
        Constraint & Fixed coordinate & Training points \\
        \midrule
        PDE interior & --- & $15{,}000$ \\
        Initial condition & $\tau=0$ & $100{,}000$ \\
        Left boundary & $x=x_{\min}$ & $100{,}000$ \\
        Right boundary & $x=x_{\max}$ & $100{,}000$ \\
        \bottomrule
    \end{tabular}
\end{table}

Training and validation panels are generated independently. Within each seed, all controlled architectures use identical interior-collocation, initial-condition, log-moneyness-boundary, and validation points. No separate variance-boundary training or validation panel is used.

Predictive evaluation uses two independently generated and frozen $10{,}000$-point reference panels sampled uniformly over the same evaluation domain using different random seeds. The first panel is used for architecture development and the controlled studies reported in Appendix~\ref{app:controlled_ablations}. The second panel is held out from development and is used for the frozen confirmatory benchmark reported in the main text and Appendix ~\ref{app:final_benchmark}.

Reference prices are evaluated from the semi-analytic Heston representation using numerical integration over $[10^{-6},500]$ with $10{,}000$ integration points. Reference sensitivities are obtained by automatic differentiation through the differentiable pricing implementation, using $20{,}000$ integration points for the Greek reference calculations. The same frozen reference values are reused for every architecture and seed. The oracle is used exclusively for predictive evaluation and never for optimization, early stopping, or checkpoint selection.

\subsection{Semi-analytic reference oracle.}
Reference prices are computed from the semi-analytic Heston representation, with the probability integrals evaluated by the trapezoidal rule on a uniform $10{,}000$-point grid over $z\in[10^{-6},500]$ in 64-bit precision. Reference $\Delta$, $K\Gamma$, and $u_v$ are obtained by automatic differentiation of the same differentiable pricing implementation, using a $20{,}000$-point integration grid and nested differentiation for the second derivative entering $K\Gamma$.

We verified numerical convergence independently of the learned models. On a deterministic $200$-point subset of the final evaluation panel, increasing the quadrature resolution to $40{,}000$--$160{,}000$ points changed all reported reference quantities by at most $2.3\times10^{-8}$. A full-panel check over all $10{,}000$ evaluation points, additionally extending the integration interval to $[10^{-9},1000]$, produced maximum differences below $1.9\times10^{-5}$ and mean differences below $2.3\times10^{-6}$. These discrepancies correspond to at most $0.016\%$ of the smallest model MAE and $0.22\%$ of the smallest difference between model means, leaving the reported model comparisons unchanged.

As an implementation-level cross-check, we additionally evaluated a deterministic stratified 200-point subset using an independently implemented Heston characteristic-function solver with adaptive quadrature. The corresponding $\Delta$, $K\Gamma$, and $u_v$ values were obtained using independent high-order finite differences. Maximum absolute discrepancies relative to the project oracle were $8.3\times10^{-6}$ for $u$, $2.1\times10^{-6}$ for $\Delta$, $1.7\times10^{-5}$ for $K\Gamma$, and $1.1\times10^{-6}$ for $u_v$. These differences are negligible relative to the predictive errors and between-model differences reported in this work and do not affect the resulting model comparisons.

\subsection{Final Model Specifications and Capacity Control}
\label{app:model_specifications}

The final benchmark compares architectures of similar trainable size while preserving their native structural geometry. Parameter counts are verified directly from the instantiated model parameter trees. The resulting configurations are summarized in Table~\ref{tab:model_capacity}.

\begin{table*}[t]
\centering
\caption{
Architecture and capacity specification for the controlled models. Parameter counts are verified from the instantiated parameter trees. Baselines preserve their native architectural structure while remaining close to the final \betinn{} parameter count.
}
\label{tab:model_capacity}
\resizebox{\textwidth}{!}{%
\begin{tabular}{lllllrr}
\toprule
Method
& Hidden / branch geometry
& $q$
& Conditioning / residual
& Params
& Mismatch (\%) \\
\midrule

\betinn{}
& $(20,20)$
& $4$
& conditioner $(12,10)$; joint residual $16$
& $3{,}595$
& $0.00$ \\

TINN
& $(37,37)$
& $1$
& temporal conditioner $(10,10)$
& $3{,}663$
& $1.89$ \\

PINN
& $(55,55)$
& ---
& ---
& $3{,}631$
& $1.00$ \\

PI-DeepONet
& branch/trunk/latent $39/39/39$
& ---
& symmetric $1{:}1{:}1$ geometry
& $3{,}511$
& $2.34$ \\

JointResidual (TINN)
& $(34,38)$
& $1$
& temporal conditioner $(7,9)$; joint residual $16$
& $3{,}577$
& $0.50$ \\

\bottomrule
\end{tabular}%
}
\end{table*}

Capacity control removes large differences in raw parameter count while preserving the defining geometry of each architecture. It helps distinguish architectural differences from simple increases in model size, while the compared function classes remain distinct. All parameter counts reported are computed directly from the instantiated parameter trees, including weights, biases, conditioner parameters, modulation tensors, and residual-pathway parameters where applicable.

The final \betinn{} configuration uses $q=4$ adaptation directions and conditioning vector
\[
c=(\tau,r,\rho,\kappa,\theta,\sigma).
\]
The structured backbone receives
\[
\zeta=(x,v,r,\rho,\kappa,\theta,\sigma),
\]
and uses hidden widths $(20,20)$. The conditioner uses hidden widths $(12,10)$, while the joint residual pathway receives the complete state--time--parameter input
\[
y=(\tau,x,v,r,\rho,\kappa,\theta,\sigma)
\]
through a single hidden layer of width $16$. The backbone contains five independently modulated parameter groups, so with $q=4$ the conditioner produces $20$ modulation coefficients. The resulting model contains $3{,}595$ trainable parameters. 

The joint-pathway output weights are initialized at $10^{-3}$ of the standard initialization scale and its output bias is initialized to zero, keeping the initial residual correction small while preserving nonzero gradients. For $q>1$, adaptation-direction initialization is scaled by $1/\sqrt{q}$ where required to preserve the corresponding $q=1$ initialization scale.

As an implementation check, the generalized $q=1$ modulation was verified against the reference time-induced formulation for both forward predictions and the derivatives required by the Heston residual.

\paragraph{TINN configuration.}
The parametric TINN baseline retains the temporal-only modulation of the original construction, with $q=1$ and conditioning variable $c=\tau$. Its structured backbone receives
\[
\zeta=(x,v,r,\rho,\kappa,\theta,\sigma),
\]
so that the Heston parameters enter as conventional backbone inputs but do not control the effective parameter modulation. The backbone uses hidden widths $(37,37)$ and the temporal conditioner uses hidden widths $(10,10)$, yielding 3,663 trainable parameters. The original fixed-parameter TINN is recovered when $\mu$ is held constant.

\paragraph{PI-DeepONet configuration.}
The controlled physics-informed DeepONet uses the Heston parameter vector $\mu=(r,\rho,\kappa,\theta,\sigma)$ as branch input and the state--time coordinates $(\tau,x,v)$ as trunk input. The branch and trunk networks have architectures
\[
5\rightarrow 39\rightarrow 39
\qquad\text{and}\qquad
3\rightarrow 39\rightarrow 39,
\]
respectively, with a $\tanh$ hidden layer and a linear 39-dimensional latent projection. Their latent representations are combined through an inner product plus a learned scalar bias,
\[
u(\tau,x,v,\mu)
=
\sum_{j=1}^{39}
b_j(\mu)\,t_j(\tau,x,v)
+
b_0.
\]
The resulting model contains 3,511 trainable parameters.

\subsection{Optimization and Checkpoint Selection}
\label{app:optimization_details}

All controlled architectures are implemented and optimized in $64$-bit floating-point precision. Training uses the same Levenberg--Marquardt \citep{more2006levenberg} implementation applied to the complete physics-informed residual system.

Optimization is limited to $3{,}000$ steps. Validation physics loss is evaluated after every optimization step, with early stopping using a patience of $250$ steps and zero minimum improvement. For each run, the checkpoint attaining the minimum validation physics loss is restored before predictive evaluation. Neither the semi-analytic reference solution nor any derivative targets are used for checkpoint selection.

\subsection{Evaluation Metrics and Diagnostic Subsets}
\label{app:extended_evaluation_metrics}

The primary Heston evaluation metric is MAE. Extended global evaluation additionally reports RMSE, while regional and structural diagnostics characterize localized and constraint-related behavior.

Derivative evaluation includes $\Delta$, strike-scaled Gamma $K\Gamma$, variance sensitivity $u_v$, spot-volatility sensitivity $2\sqrt{v}\,u_v$, normalized calendar Theta $-u_\tau$, and normalized interest-rate sensitivity $u_r$. All derivatives are obtained by automatic differentiation of the trained surrogate.

Regional diagnostics use fixed subsets defined independently of model
predictions. Table~\ref{tab:regional_masks} reports the exact definitions and
the number of points selected from the frozen $10{,}000$-point evaluation
panel. The masks are fixed before model-level aggregation and may overlap.

\begin{table}[t]
\centering
\small
\caption{
Regional diagnostic subsets.
Definitions and numbers of selected points on the frozen
$10{,}000$-point evaluation panel.
}
\label{tab:regional_masks}
\begin{tabular}{p{0.43\linewidth}p{0.39\linewidth}r}
\toprule
Subset & Definition & Points \\
\midrule
ATM tight
& $|x|\le 0.05$
& 253 \\

ATM
& $|x|\le 0.10$
& 501 \\

ATM wide
& $|x|\le 0.25$
& 1297 \\

Very near maturity
& $\tau\le 0.05$
& 109 \\

Near maturity
& $\tau\le 0.10$
& 204 \\

Short maturity
& $\tau\le 0.25$
& 520 \\

Long maturity
& $\tau\ge 4.5$
& 996 \\

ATM \& near maturity
& $|x|\le 0.10,\ \tau\le 0.10$
& 8 \\

ATM wide \& short maturity
& $|x|\le 0.25,\ \tau\le 0.25$
& 62 \\

$x>1.8$
& $x>1.8$
& 443 \\

$x>1.9$
& $x>1.9$
& 224 \\

High-$x$, high-$\tau$, high-$\theta$
& $x\ge1.6,\ \tau\ge3.75,\ \theta\ge0.7525$
& 48 \\

Near $v_{\min}$
& $v\le0.145$
& 520 \\

Variance interior
& $0.325\le v\le0.775$
& 4885 \\

Near $v_{\max}$
& $v\ge0.955$
& 554 \\

Feller satisfied
& $2\kappa\theta\ge\sigma^2$
& 8770 \\

Feller violated
& $2\kappa\theta<\sigma^2$
& 1230 \\
\bottomrule
\end{tabular}
\end{table}

We also distinguish parameter configurations satisfying or violating the Feller condition
\[
2\kappa\theta\geq\sigma^2.
\]

Shape and sign diagnostics use a tolerance $\varepsilon=10^{-4}$ and record
violations of
\[
\Delta<-\varepsilon,\qquad
\Delta>1+\varepsilon,\qquad
K\Gamma<-\varepsilon,\qquad
u_v<-\varepsilon,\qquad
-u_\tau>\varepsilon,\qquad
u_r<-\varepsilon.
\]
The same tolerance applied to the frozen semi-analytic oracle yields zero
violations for all six conditions. All regional and structural masks are fixed before model-level aggregation.

\subsection{Compute and Implementation}
\label{app:compute_implementation}

Each experiment used a single NVIDIA A100 80\,GB PCIe GPU with CUDA~12.4. All controlled models were implemented in JAX/Flax with 64-bit computation enabled. Optimization uses 64-bit LM updates, and all restored checkpoints and predictive evaluations are represented in 64-bit floating-point precision. Parameter counts were obtained from instantiated parameter trees, and automatic differentiation was used for both the physics-informed training residuals and derivative evaluation.

Table~\ref{tab:computational_efficiency} reports the computational profile of the final controlled models. Training wall-clock times and selected optimization steps were recovered from the corresponding final training runs, whereas inference was reprofiled from the frozen best checkpoints under a common single-GPU protocol. JAX compilation was measured separately and excluded from steady-state timings; compilation times ranged from approximately $0.15$--$0.39$\,s for $B=1$ and $0.67$--$0.97$\,s for the larger batches. All timed forward calls were synchronized before measurement.

Across the four architectures, single-point forward latency ranges from $0.10$ to $0.17$\,ms, while batches of $10{,}000$ points achieve approximately $4.3\times10^7$--$7.3\times10^7$ samples/s. \betinn{} incurs the largest inference cost among the controlled models under this forward-pass benchmark, consistent with the additional computation introduced by its conditional parameter modulation and joint residual pathway. Its recorded training wall-clock time is higher than those of TINN and PINN but lower than that of PI-DeepONet.

\begin{table}[t]
\centering
\caption{
Computational profiling of the final models. Training time and the optimization step attaining the minimum validation physics loss are recovered from the final training runs and reported as mean $\pm$ std over six seeds. Inference is profiled post hoc from the restored best-validation checkpoints. For $B=1$, latency is the median forward-pass time; for larger batches, throughput is the batch size divided by the median forward-pass time. Peak allocation denotes the maximum JAX-managed device allocation during forward inference up to
$B=10{,}000$.
}
\label{tab:computational_efficiency}
\resizebox{\linewidth}{!}{%
\begin{tabular}{lrcccccc}
\toprule
Model
& Params.
& Train time (s)
& Best val.\ step
& $B=1$ latency (ms)
& $B=1000$ throughput
& $B=10000$ throughput
& Peak alloc. (MiB) \\
\midrule
\betinn{}
& 3595
& $432 \pm 26$
& $72 \pm 21$
& $0.169 \pm 0.005$
& $(5.00 \pm 0.14)\times10^6$
& $(4.26 \pm 0.08)\times10^7$
& 256.7 \\

TINN
& 3663
& $263 \pm 7$
& $74 \pm 10$
& $0.126 \pm 0.004$
& $(6.76 \pm 0.08)\times10^6$
& $(5.30 \pm 0.07)\times10^7$
& 160.7 \\

PINN
& 3631
& $226 \pm 4$
& $67 \pm 5$
& $0.101 \pm 0.002$
& $(1.07 \pm 0.02)\times10^7$
& $(7.27 \pm 0.05)\times10^7$
& 80.9 \\

PI-DeepONet
& 3511
& $643 \pm 84$
& $770 \pm 135$
& $0.107 \pm 0.003$
& $(8.39 \pm 0.19)\times10^6$
& $(6.31 \pm 0.09)\times10^7$
& 99.8 \\
\bottomrule
\end{tabular}%
}
\end{table}

\section{Controlled Architecture and Capacity Studies}
\label{app:controlled_ablations}

This section studies the architectural and capacity choices underlying \betinn{}. All experiments reported here use the common Heston training and evaluation protocol described in Appendix~\ref{app:experimental_configuration} and are evaluated from the restored best validation checkpoint.

Unless stated otherwise, results are computed over the same six development seeds and are reported as mean $\pm$ standard deviation. These experiments are used for architecture selection and diagnosis; the frozen final benchmark is evaluated separately on the confirmatory seeds reported in the main text.

We first vary the dimensionality of the conditional modulation space, then study backbone capacity at fixed modulation dimensionality, compare the resulting configurations against independently scaled controls, and finally examine the contributions of the principal architectural components.

\subsection{Modulation-Dimensionality Ablation}
\label{app:q_ablation}

We first vary the number of adaptation directions $q$ while fixing the structured backbone to $(20,20)$, the conditioner to $(12,10)$, and retaining both parameter conditioning and the joint residual pathway. This experiment varies modulation dimensionality at fixed backbone geometry, but does not hold total parameter count constant: increasing $q$ introduces additional modulation parameters.

\begin{table}[t]
\centering
\caption{
Modulation-dimensionality ablation with fixed $(20,20)$ structured backbone
and $(12,10)$ conditioner. Values are mean $\pm$ std over six development
seeds. Lower is better.
}
\label{tab:q_ablation}
\resizebox{\linewidth}{!}{%
\begin{tabular}{lccccc}
\toprule
Configuration
& Params
& Price MAE $\downarrow$
& $\Delta$ MAE $\downarrow$
& $K\Gamma$ MAE $\downarrow$
& $u_v$ MAE $\downarrow$ \\
\midrule

$q=1$
& 1,630
& $0.00769 \pm 0.00110$
& $0.01849 \pm 0.00179$
& $0.15218 \pm 0.00999$
& $0.00747 \pm 0.00132$ \\

$q=2$
& 2,285
& $0.00777 \pm 0.00084$
& $0.01864 \pm 0.00103$
& $0.15013 \pm 0.00562$
& $0.00736 \pm 0.00078$ \\

$q=3$
& 2,940
& $0.00799 \pm 0.00088$
& $0.01871 \pm 0.00125$
& $0.14326 \pm 0.00687$
& $0.00753 \pm 0.00086$ \\

$q=4$
& 3,595
& $\mathbf{0.00625 \pm 0.00095}$
& $\mathbf{0.01605 \pm 0.00201}$
& $\mathbf{0.13615 \pm 0.00919}$
& $\mathbf{0.00582 \pm 0.00116}$ \\

$q=5$
& 4,250
& $0.00736 \pm 0.00090$
& $0.01771 \pm 0.00151$
& $0.14396 \pm 0.00673$
& $0.00704 \pm 0.00079$ \\

$q=6$
& 4,905
& $0.00696 \pm 0.00124$
& $0.01726 \pm 0.00165$
& $0.14460 \pm 0.01120$
& $0.00649 \pm 0.00116$ \\

\bottomrule
\end{tabular}%
}
\end{table}

Table~\ref{tab:q_ablation} shows that the $q=1$, $q=2$, and $q=3$ configurations yield broadly similar errors, while $q=4$ produces a clear improvement across all four primary quantities. Increasing the modulation dimensionality further to $q=5$ or $q=6$ does not improve this, both configurations have higher errors despite their larger parameter counts.

Thus, $q=4$ provides the strongest development performance among the modulation dimensions evaluated at fixed backbone geometry. Because parameter count increases with $q$, the modulation sweep alone does not isolate dimensionality from model capacity; the following capacity-controlled studies address this distinction.

\subsection{Backbone Capacity at Fixed $q=4$}
\label{app:capacity_ablation}

We next fix $q=4$ and vary the capacity of the structured backbone. The clean capacity sweep keeps the conditioner fixed at $(12,10)$ and uses backbones $(12,12)$, $(16,16)$, and $(20,20)$, corresponding to $1{,}915$, $2{,}675$, and $3{,}595$ trainable parameters, respectively.

For reference, we also include a $1{,}385$-parameter diagnostic using a $(10,10)$ backbone. Meeting this restrictive budget additionally requires reducing the conditioner to $(6,6)$, so this point is not part of the clean fixed-conditioner capacity curve.

\begin{table}[t]
\centering
\caption{
Backbone-capacity ablation at fixed modulation dimensionality $q=4$. Values are mean $\pm$ std over six development seeds. The $1{,}385$-parameter configuration is a fixed-budget diagnostic that also uses a smaller $(6,6)$ conditioner; the clean fixed-conditioner capacity sweep spans $1{,}915$--$3{,}595$ parameters. Lower is better.
}
\label{tab:capacity_ablation}
\resizebox{\linewidth}{!}{%
\begin{tabular}{lccccc}
\toprule
Backbone
& Params
& Price MAE $\downarrow$
& $\Delta$ MAE $\downarrow$
& $K\Gamma$ MAE $\downarrow$
& $u_v$ MAE $\downarrow$ \\
\midrule

$(10,10)$, fixed-budget diagnostic
& 1,385
& $0.00845 \pm 0.00098$
& $0.01908 \pm 0.00103$
& $0.13680 \pm 0.00795$
& $0.00800 \pm 0.00095$ \\

$(12,12)$
& 1,915
& $0.00758 \pm 0.00110$
& $0.01776 \pm 0.00150$
& $0.13925 \pm 0.00589$
& $0.00737 \pm 0.00098$ \\

$(16,16)$
& 2,675
& $0.00687 \pm 0.00175$
& $0.01650 \pm 0.00261$
& $\mathbf{0.13192 \pm 0.00451}$
& $0.00636 \pm 0.00181$ \\

$(20,20)$
& 3,595
& $\mathbf{0.00625 \pm 0.00095}$
& $\mathbf{0.01605 \pm 0.00201}$
& $0.13615 \pm 0.00919$
& $\mathbf{0.00582 \pm 0.00116}$ \\

\bottomrule
\end{tabular}%
}
\end{table}

Table~\ref{tab:capacity_ablation} shows that increasing backbone capacity consistently improves Price, $\Delta$, and $u_v$ over the clean capacity sweep. The behavior of $K\Gamma$ is less monotonic: the $(16,16)$ configuration attains the lowest mean second-order error, whereas the larger $(20,20)$ model provides the strongest overall development profile across Price, $\Delta$, and $u_v$ while retaining competitive $K\Gamma$ accuracy.

The results indicate that the richer $q=4$ conditional parameterization benefits from sufficient backbone capacity and motivate evaluating the selected models against independently scaled architectural controls.

\subsection{Capacity-Controlled Comparisons}
\label{app:capacity_matched}

The preceding experiments show that increased capacity benefits the structured model, but do not establish whether similar gains can be reproduced simply by enlarging competing architectures. Here, we scale TINN, PINN, and the JointResidual control independently around two \betinn{} operating points while preserving the native geometry of each architecture.

Table~\ref{tab:matched_2675} reports the lower-capacity comparison centered on the $2{,}675$-parameter $(16,16)$ \betinn{} configuration, with controls lying within approximately $2\%$ of the same parameter budget. Table~\ref{tab:matched_3595} repeats this comparison around the selected $3{,}595$-parameter configuration using independently scaled controls.

\begin{table}[t]
\centering
\caption{
Capacity-controlled development comparison around the $2{,}675$-parameter \betinn{} configuration. Values are mean $\pm$ std over six development seeds. Controls are independently scaled while preserving their native architectural geometry. Lower is better.}
\label{tab:matched_2675}
\resizebox{\linewidth}{!}{%
\begin{tabular}{lccccc}
\toprule
Method
& Params
& Price MAE $\downarrow$
& $\Delta$ MAE $\downarrow$
& $K\Gamma$ MAE $\downarrow$
& $u_v$ MAE $\downarrow$ \\
\midrule

TINN
& 2,727
& $0.00794 \pm 0.00105$
& $0.0181 \pm 0.0009$
& $0.1513 \pm 0.0039$
& $0.00770 \pm 0.00118$ \\

PINN
& 2,623
& $0.00847 \pm 0.00100$
& $0.0190 \pm 0.0011$
& $0.1475 \pm 0.0045$
& $0.00801 \pm 0.00086$ \\

JointResidual
& 2,665
& $0.00797 \pm 0.00168$
& $0.0182 \pm 0.0021$
& $0.1482 \pm 0.0103$
& $0.00751 \pm 0.00151$ \\

\betinn{} $(16{\times}16)$
& 2,675
& $\mathbf{0.00687 \pm 0.00175}$
& $\mathbf{0.0165 \pm 0.0026}$
& $\mathbf{0.1319 \pm 0.0045}$
& $\mathbf{0.00636 \pm 0.00181}$ \\

\bottomrule
\end{tabular}%
}
\end{table}

As shown in Table~\ref{tab:matched_2675}, \betinn{} attains the lowest mean error for all four primary quantities at the lower-capacity operating point. This comparison indicates that the advantage of the structured conditional representation is already present below the final model scale.

\begin{table}[t]
\centering
\caption{
Capacity-controlled development comparison at the final \betinn{} scale. Values are mean $\pm$ std over six development seeds. This table reports development evidence and is distinct from the frozen final benchmark.
}
\label{tab:matched_3595}
\resizebox{\linewidth}{!}{%
\begin{tabular}{lccccc}
\toprule
Method
& Params
& Price MAE $\downarrow$
& $\Delta$ MAE $\downarrow$
& $K\Gamma$ MAE $\downarrow$
& $u_v$ MAE $\downarrow$ \\
\midrule

TINN
& 3,663
& $0.00759 \pm 0.00050$
& $0.0179 \pm 0.0009$
& $0.148 \pm 0.008$
& $0.00725 \pm 0.00067$ \\

PINN
& 3,631
& $0.00859 \pm 0.00122$
& $0.0187 \pm 0.0014$
& $0.149 \pm 0.007$
& $0.00816 \pm 0.00116$ \\

JointResidual
& 3,577
& $0.00774 \pm 0.00146$
& $0.0182 \pm 0.0020$
& $0.151 \pm 0.009$
& $0.00741 \pm 0.00148$ \\

\betinn{}
& 3,595
& $\mathbf{0.00625 \pm 0.00095}$
& $\mathbf{0.0161 \pm 0.0020}$
& $\mathbf{0.136 \pm 0.009}$
& $\mathbf{0.00582 \pm 0.00116}$ \\

\bottomrule
\end{tabular}%
}
\end{table}

Table~\ref{tab:matched_3595} yields the same qualitative conclusion. \betinn{} attains lower mean error than the independently scaled TINN, PINN, and JointResidual controls on all four primary metrics. The development advantage therefore persists after competing architectures are enlarged to comparable capacity.

\subsection{Capacity Architectural Component Ablations}
\label{app:results_ablations}

We next isolate the architectural ingredients of \betinn{} near the final model scale. Each controlled variant preserves its native architectural semantics and is scaled approximately to the $3{,}595$-parameter \betinn{} configuration. 

\begin{table*}[t]
\centering
\caption{
Capacity architectural component ablations on the six development seeds. Controlled variants are scaled approximately to the final \betinn{} parameter budget while preserving their architectural structure. Values are mean
$\pm$ std. Lower is better.
}
\label{tab:component_final_capacity}
\resizebox{\textwidth}{!}{%
\begin{tabular}{lrrrrrr}
\toprule
Configuration
& Params
& Mismatch (\%)
& Price MAE
& $\Delta$ MAE
& $K\Gamma$ MAE
& $u_v$ MAE \\
\midrule

\betinn{} (reference)
& 3595 & 0.00
& $\mathbf{0.00625 \pm 0.00095}$
& $\mathbf{0.01605 \pm 0.00201}$
& $\mathbf{0.1361 \pm 0.0092}$
& $\mathbf{0.00582 \pm 0.00116}$ \\

TINN
& 3663 & 1.89
& $0.00759 \pm 0.00050$
& $0.01788 \pm 0.00092$
& $0.1479 \pm 0.0077$
& $0.00725 \pm 0.00067$ \\

TINN + parameter conditioning
& 3595 & 0.00
& $0.00823 \pm 0.00147$
& $0.01857 \pm 0.00229$
& $0.1491 \pm 0.0109$
& $0.00774 \pm 0.00142$ \\

TINN + modified-MLP gating
& 3625 & 0.83
& $0.00785 \pm 0.00075$
& $0.01809 \pm 0.00053$
& $0.1605 \pm 0.0066$
& $0.00764 \pm 0.00062$ \\

TINN + highway gating
& 3695 & 2.78
& $0.00812 \pm 0.00108$
& $0.01842 \pm 0.00128$
& $0.1499 \pm 0.0101$
& $0.00783 \pm 0.00104$ \\

TINN + adaptive residual
& 3665 & 1.95
& $0.00799 \pm 0.00078$
& $0.01869 \pm 0.00106$
& $0.1596 \pm 0.0066$
& $0.00795 \pm 0.00074$ \\

JointResidual
& 3577 & 0.50
& $0.00774 \pm 0.00146$
& $0.01822 \pm 0.00198$
& $0.1513 \pm 0.0094$
& $0.00741 \pm 0.00148$ \\

TINN + Fourier features
& 3545 & 1.39
& $0.00788 \pm 0.00069$
& $0.01828 \pm 0.00078$
& $0.1765 \pm 0.0054$
& $0.00743 \pm 0.00065$ \\

TINN + joint residual + $q=4$
& 3595 & 0.00
& $0.00832 \pm 0.00090$
& $0.01900 \pm 0.00112$
& $0.1451 \pm 0.0038$
& $0.00786 \pm 0.00086$ \\

TINN + joint residual + conditioning
& 3641 & 1.28
& $0.00793 \pm 0.00145$
& $0.01806 \pm 0.00192$
& $0.1524 \pm 0.0109$
& $0.00735 \pm 0.00134$ \\

\betinn{} w/o joint residual
& 3434
& 4.48
& $0.00764 \pm 0.00130$
& $0.01794 \pm 0.00235$
& $0.1382 \pm 0.0078$
& $0.00707 \pm 0.00146$
\\

\bottomrule
\end{tabular}%
}
\end{table*}

Table~\ref{tab:component_final_capacity} shows that none of the tested single architectural modifications to a base TINN reproduces the final \betinn{} performance. Parameter conditioning, alternative gating mechanisms, a direct joint residual pathway, and Fourier features all produce distinct error profiles, but none matches the joint performance of the complete \betinn{} representation across the four primary quantities.

The compositional controls in Table~\ref{tab:component_final_capacity} show that combining the joint residual pathway with either $q=4$ modulation or parameter conditioning alone does not recover the full architecture. Conversely, removing the joint residual pathway from the $q=4$ parameter-conditioned model increases the mean Price, $\Delta$, and $u_v$ errors, with the full \betinn{} configuration performing better in five of the six paired development seeds for each quantity. The effect on $K\Gamma$ is smaller and evenly split across seeds. Taken together, these results support complementary roles for multi-directional modulation, parameter-conditioned adaptation, and the joint pathway in the joint value–derivative performance of the full \betinn{} configuration.

To complement the aggregate means in Table~\ref{tab:component_final_capacity}, Table~\ref{tab:component_paired} reports the corresponding seed-wise paired effects relative to \betinn{}. These paired comparisons show that the aggregate advantage of \betinn{} is also reflected across the common development seeds and not just being driven by a single run.

We further examined whether the final \betinn{} checkpoints materially use the governing-parameter dependence of the conditioner. Although the learned $q=4$ coefficient trajectories are strongly dominated by a single time-associated principal direction, their  parameter-dependent component is functionally consequential. Freezing only the governing-parameter inputs to the conditioner at the centers of their training ranges, while preserving the true parameters in the backbone and joint residual pathway, increases Price, $\Delta$, $K\Gamma$, and $u_v$ error in all six confirmatory seeds. Thus, the learned modulation is geometrically close to low-dimensional in coefficient space, while still making material use of parameter-dependent adaptation in the trained surrogate.

\subsection{Selection of the Final \betinn{} Configuration}
\label{app:betinn_selection}

\begin{table*}[t]
\centering
\caption{
Paired effects of final-capacity architectural ablations relative to \betinn{} over the six common development seeds. Differences are computed seed-wise as $\mathrm{ablation}-\betinn{}$, so positive values indicate larger error for the ablation. Parentheses report the number of seeds on which the ablation attains lower error than \betinn{}.
}
\label{tab:component_paired}
\resizebox{\textwidth}{!}{%
\begin{tabular}{lrrrrr}
\toprule
Configuration & Params & Price diff. & $\Delta$ diff.
& $K\Gamma$ diff. & $u_v$ diff. \\
\midrule

TINN
& 3663 & $+0.00133$ (1/6) & $+0.00182$ (2/6)
& $+0.0118$ (2/6) & $+0.00143$ (1/6) \\

TINN + parameter conditioning
& 3595 & $+0.00197$ (0/6) & $+0.00251$ (0/6)
& $+0.0129$ (0/6) & $+0.00192$ (0/6) \\

TINN + modified-MLP gating
& 3625 & $+0.00159$ (1/6) & $+0.00204$ (1/6)
& $+0.0243$ (0/6) & $+0.00182$ (0/6) \\

TINN + highway gating
& 3695 & $+0.00186$ (1/6) & $+0.00237$ (1/6)
& $+0.0137$ (2/6) & $+0.00201$ (1/6) \\

TINN + adaptive residual
& 3665 & $+0.00173$ (1/6) & $+0.00264$ (1/6)
& $+0.0234$ (0/6) & $+0.00213$ (1/6) \\

JointResidual
& 3577 & $+0.00149$ (1/6) & $+0.00217$ (1/6)
& $+0.0151$ (1/6) & $+0.00160$ (1/6) \\

TINN + Fourier features
& 3545 & $+0.00163$ (0/6) & $+0.00222$ (1/6)
& $+0.0403$ (0/6) & $+0.00162$ (0/6) \\

TINN + joint residual + $q=4$
& 3595 & $+0.00207$ (0/6) & $+0.00295$ (0/6)
& $+0.00892$ (1/6) & $+0.00204$ (0/6) \\

TINN + joint residual + conditioning
& 3641 & $+0.00167$ (0/6) & $+0.00200$ (0/6)
& $+0.0162$ (0/6) & $+0.00153$ (0/6) \\

\betinn{} w/o joint residual
& 3434
& $+0.00139$ (1/6)
& $+0.00189$ (1/6)
& $+0.00208$ (3/6)
& $+0.00125$ (1/6)
\\

\bottomrule
\end{tabular}%
}
\end{table*}

Taken together, the modulation study in Table~\ref{tab:q_ablation}, the capacity study in Table~\ref{tab:capacity_ablation}, the capacity-controlled comparisons in Tables~\ref{tab:matched_2675} and~\ref{tab:matched_3595}, and the component analyses in Tables~\ref{tab:component_final_capacity} and~\ref{tab:component_paired} support three observations.

First, multi-directional modulation benefits from sufficient backbone capacity. Second, $q=4$ provides the strongest overall development performance once that capacity is available. Third, independently enlarging TINN, PINN, and JointResidual controls does not reproduce the same joint value--derivative accuracy profile.

Based exclusively on these development experiments, we select
\[
q=4,\qquad
\text{backbone}=(20,20),\qquad
\text{conditioner}=(12,10)
\]
as the final \betinn{} configuration, which produces $3{,}595$ trainable parameters. The architecture and its training, checkpoint-selection, and evaluation protocols are then frozen before evaluation on the confirmatory seeds used in the main benchmark.

The $(16,16)$, $2{,}675$-parameter configuration is retained only as a secondary capacity--accuracy operating point and is not used to modify the final confirmatory model.

\section{Extended Results for the Final Benchmark}
\label{app:final_benchmark}

This section extends the aggregate comparison reported in the main text. Sections~\ref{app:final_seed_robustness}--\ref{app:regional_results} analyze the frozen final models using the six common confirmatory seeds. Section~\ref{app:additional_baselines} reports additional baseline architecture analyses conducted on the development seeds, while Section~\ref{app:deepsvm} evaluates the separately trained DeepSVM-adapted reference. We examine seed-wise robustness, extended metrics and sensitivities, validation-loss associations, and regional performance.

\subsection{Seed-Wise Robustness}
\label{app:final_seed_robustness}

The main benchmark reports averages across the six final seeds. Table~\ref{tab:final_paired} complements these aggregate results with paired seed-wise differences between \betinn{} and each main baseline. Because all models are evaluated on common seeds and the same frozen evaluation panel, the comparison directly reveals how consistently the aggregate advantage appears across independent training runs.

\begin{table}[t]
\centering
\caption{
Paired final-benchmark differences between \betinn{} and each main baseline over the six common confirmatory seeds. Differences are computed seed-wise as $\betinn{}-\mathrm{baseline}$, so negative values indicate lower \betinn{} error. ``Wins'' denotes the number of seeds on which \betinn{} attains lower
error.
}
\label{tab:final_paired}
\resizebox{0.6\linewidth}{!}{%
\begin{tabular}{llrrr}
\toprule
Baseline & Metric & Mean diff. & Median diff. & \betinn{} wins \\
\midrule

\multirow{4}{*}{TINN}
& Price       & $-0.00164$  & $-0.00118$  & 6/6 \\
& $\Delta$    & $-0.00177$  & $-0.00159$  & 5/6 \\
& $K\Gamma$   & $-0.0149$   & $-0.0157$   & 6/6 \\
& $u_v$       & $-0.00141$  & $-0.00111$  & 6/6 \\
\midrule

\multirow{4}{*}{PINN}
& Price       & $-0.00110$  & $-0.000959$ & 5/6 \\
& $\Delta$    & $-0.000783$ & $-0.000504$ & 4/6 \\
& $K\Gamma$   & $-0.00601$  & $-0.00768$  & 4/6 \\
& $u_v$       & $-0.000880$ & $-0.000587$ & 5/6 \\
\midrule

\multirow{4}{*}{PI-DeepONet}
& Price       & $-0.00228$  & $-0.00277$  & 5/6 \\
& $\Delta$    & $-0.00521$  & $-0.00532$  & 6/6 \\
& $K\Gamma$   & $-0.0562$   & $-0.0578$   & 6/6 \\
& $u_v$       & $-0.00382$  & $-0.00391$  & 6/6 \\

\bottomrule
\end{tabular}%
}
\end{table}

As shown in Table~\ref{tab:final_paired}, the improvement is not driven by a
single training run. Relative to TINN, \betinn{} attains lower error on all six
seeds for Price, $K\Gamma$, and $u_v$, and on five of six seeds for $\Delta$.
The comparison with PINN is less uniform for the derivative metrics but still
favors \betinn{} on four to five of the six common seeds. Against PI-DeepONet,
\betinn{} wins at least five of six seeds for every primary quantity.

\subsection{Extended Global Accuracy and Sensitivities}
\label{app:extended_global_accuracy}

The primary benchmark focuses on MAE for Price, $\Delta$, $K\Gamma$, and
$u_v$. Table~\ref{tab:global_metrics} extends this evaluation in two directions:
it additionally reports RMSE, and it evaluates spot-volatility sensitivity
$2\sqrt{v}\,u_v$, normalized calendar Theta $-u_\tau$, and interest-rate
sensitivity $u_r$.

\begin{table}[t]
\centering
\caption{
Global pricing and sensitivity errors on the frozen final evaluation panel. Values are mean $\pm$ std over six seeds. Lower is better.
}
\label{tab:global_metrics}
\resizebox{\linewidth}{!}{%
\begin{tabular}{llccccccc}
\toprule
Metric & Method
& Price
& $\Delta$
& $K\Gamma$
& $u_v$
& $2\sqrt{v}\,u_v$
& $-u_\tau$
& $u_r$ \\
\midrule

\multirow{4}{*}{MAE}
& \betinn{}
& $0.00746 \pm 0.00097$
& $0.0180 \pm 0.0010$
& $0.143 \pm 0.005$
& $0.00712 \pm 0.00089$
& $0.00942 \pm 0.00114$
& $0.00475 \pm 0.00063$
& $0.0597 \pm 0.0151$ \\

& TINN
& $0.00910 \pm 0.00062$
& $0.0197 \pm 0.0009$
& $0.158 \pm 0.006$
& $0.00853 \pm 0.00046$
& $0.0113 \pm 0.0006$
& $0.00566 \pm 0.00027$
& $0.0808 \pm 0.0118$ \\

& PINN
& $0.00857 \pm 0.00054$
& $0.0188 \pm 0.0005$
& $0.149 \pm 0.003$
& $0.00800 \pm 0.00049$
& $0.0106 \pm 0.0007$
& $0.00528 \pm 0.00030$
& $0.0817 \pm 0.0090$ \\

& PI-DeepONet
& $0.00974 \pm 0.00063$
& $0.0232 \pm 0.0007$
& $0.199 \pm 0.016$
& $0.0109 \pm 0.0005$
& $0.0145 \pm 0.0007$
& $0.00709 \pm 0.00028$
& $0.0774 \pm 0.0076$ \\

\midrule

\multirow{4}{*}{RMSE}
& \betinn{}
& $0.0124 \pm 0.0011$
& $0.0382 \pm 0.0009$
& $0.397 \pm 0.022$
& $0.0123 \pm 0.0014$
& $0.0148 \pm 0.0017$
& $0.00906 \pm 0.00150$
& $0.106 \pm 0.027$ \\

& TINN
& $0.0148 \pm 0.0009$
& $0.0412 \pm 0.0014$
& $0.423 \pm 0.021$
& $0.0144 \pm 0.0006$
& $0.0174 \pm 0.0008$
& $0.0101 \pm 0.0008$
& $0.155 \pm 0.025$ \\

& PINN
& $0.0141 \pm 0.0009$
& $0.0390 \pm 0.0009$
& $\mathbf{0.385 \pm 0.010}$
& $0.0138 \pm 0.0008$
& $0.0166 \pm 0.0011$
& $0.00967 \pm 0.00060$
& $0.160 \pm 0.022$ \\

& PI-DeepONet
& $0.0152 \pm 0.0010$
& $0.0450 \pm 0.0011$
& $0.502 \pm 0.053$
& $0.0180 \pm 0.0007$
& $0.0218 \pm 0.0010$
& $0.0137 \pm 0.0006$
& $0.123 \pm 0.011$ \\

\bottomrule
\end{tabular}%
}
\end{table}

Table~\ref{tab:global_metrics} shows that the primary MAE advantage extends to the additional sensitivities: \betinn{} attains the lowest mean MAE for $2\sqrt{v}\,u_v$, $-u_\tau$, and $u_r$. The RMSE results are similarly favorable across most quantities, although PINN attains the lowest mean $K\Gamma$ RMSE. Thus, the stronger aggregate performance of \betinn{} is not restricted to the four primary MAE metrics, while the $K\Gamma$ RMSE result also shows that model ordering can depend on the error functional being
evaluated.

\subsection{Validation Physics Loss and Downstream Error}
\label{app:validation_downstream}

Figure~\ref{fig:validation_downstream_proxy} in the main text shows that validation physics loss is informative of downstream predictive quality but does not uniquely determine solution or derivative accuracy. Table~\ref{tab:validation_spearman} quantifies this relationship across the $24$ final model--seed pairs.

\begin{table}[t]
\centering
\caption{
Spearman association between the best validation physics loss and downstream prediction error across the $24$ final model--seed pairs ($4$ architectures $\times$ $6$ confirmatory seeds). Positive values indicate that runs with larger validation physics loss tend to exhibit larger downstream error.
}
\label{tab:validation_spearman}
\begin{tabular}{lc}
\toprule 
Error metric & Spearman \\
\midrule
Price MAE     & 0.578 \\
$\Delta$ MAE  & 0.682 \\
$K\Gamma$ MAE & 0.617 \\
$u_v$ MAE     & 0.698 \\
\bottomrule
\end{tabular}
\end{table}

As summarized in Table~\ref{tab:validation_spearman}, all four associations are positive, with run-level Spearman coefficients between $0.578$ and $0.698$. Validation physics loss is therefore a meaningful downstream indicator, but these moderate associations do not fully preserve predictive rankings in solution and derivative space.

\subsection{Regional Error Analysis}
\label{app:regional_results}

As global averages can obscure substantial variation across the Heston domain, Tables~\ref{tab:regional_price_delta} and~\ref{tab:regional_kgamma_uv} report the primary MAE metrics on the fixed diagnostic subsets defined in Appendix~\ref{app:extended_evaluation_metrics}. All regional masks are specified independently of model predictions.

\begin{table*}[t]
\centering
\caption{
Region-wise mean MAE for Price and $\Delta$ on the frozen final evaluation
panel. The number of evaluation points in each predefined subset is shown in
parentheses. Bold indicates the lowest mean error within each region and
metric. All regions are fixed before model-level aggregation.
}
\label{tab:regional_price_delta}
\setlength{\tabcolsep}{3.5pt}
\resizebox{\textwidth}{!}{%
\begin{tabular}{lcccccccc}
\toprule
& \multicolumn{4}{c}{Price MAE}
& \multicolumn{4}{c}{$\Delta$ MAE} \\
\cmidrule(lr){2-5}
\cmidrule(lr){6-9}
Region
& \betinn{} & TINN & PINN & PI-DeepONet
& \betinn{} & TINN & PINN & PI-DeepONet \\
\midrule

ATM (501)
& \textbf{.00672} & .00934 & .00839 & .0106
& \textbf{.00992} & .0110 & .0109 & .0141 \\

ATM tight (253)
& \textbf{.00702} & .00987 & .00890 & .0112
& \textbf{.00978} & .0110 & .0109 & .0135 \\

ATM wide (1297)
& \textbf{.00700} & .00959 & .00860 & .0111
& \textbf{.00992} & .0110 & .0109 & .0139 \\

ATM wide \& short maturity (62)
& .00432 & .00417 & \textbf{.00406} & .00623
& \textbf{.0174} & .0184 & .0194 & .0306 \\

ATM \& near maturity (8)
& .00559 & \textbf{.00479} & .00507 & .00520
& \textbf{.0318} & .0332 & .0387 & .0741 \\

\midrule

Long maturity (996)
& \textbf{.0149} & .0183 & .0179 & .0168
& \textbf{.0327} & .0363 & .0348 & .0378 \\

Near maturity (204)
& \textbf{.000940} & .00103 & .000945 & .00141
& \textbf{.00549} & .00631 & .00605 & .0137 \\

Short maturity (520)
& \textbf{.00129} & .00143 & \textbf{.00129} & .00225
& \textbf{.00542} & .00644 & .00599 & .0129 \\

Very near maturity (109)
& \textbf{.000793} & .000836 & .000837 & .00129
& \textbf{.00574} & .00628 & .00659 & .0150 \\

\midrule

Feller satisfied (8770)
& \textbf{.00713} & .00844 & .00800 & .00898
& \textbf{.0168} & .0182 & .0174 & .0210 \\

Feller violated (1230)
& \textbf{.00983} & .0138 & .0126 & .0152
& \textbf{.0262} & .0306 & .0284 & .0386 \\

High-$x$, high-$\tau$, high-$\theta$ (48)
& .0462 & \textbf{.0385} & .0399 & .0437
& .0115 & .0111 & .0112 & \textbf{.0110} \\

\midrule

Near $v_{\max}$ (554)
& \textbf{.00779} & .00881 & .00849 & .00944
& \textbf{.0192} & .0198 & .0195 & .0230 \\

Near $v_{\min}$ (520)
& \textbf{.00833} & .0111 & .0103 & .0121
& \textbf{.0219} & .0248 & .0232 & .0310 \\

Variance interior (4885)
& \textbf{.00725} & .00882 & .00828 & .00936
& \textbf{.0167} & .0185 & .0175 & .0216 \\

\midrule

$x>1.8$ (443)
& \textbf{.0183} & .0191 & .0187 & .0209
& .00467 & .00463 & \textbf{.00461} & .00482 \\

$x>1.9$ (224)
& \textbf{.0188} & .0198 & .0194 & .0216
& .00462 & \textbf{.00457} & .00457 & .00475 \\

\bottomrule
\end{tabular}%
}
\end{table*}

\begin{table*}[t]
\centering
\caption{
Region-wise mean MAE for strike-scaled Gamma $K\Gamma$ and variance
sensitivity $u_v$ on the frozen final evaluation panel. The number of
evaluation points in each predefined subset is shown in parentheses. Bold
indicates the lowest mean error within each region and metric. All regions are
fixed before model-level aggregation.
}
\label{tab:regional_kgamma_uv}
\setlength{\tabcolsep}{3.5pt}
\resizebox{\textwidth}{!}{%
\begin{tabular}{lcccccccc}
\toprule
& \multicolumn{4}{c}{$K\Gamma$ MAE}
& \multicolumn{4}{c}{$u_v$ MAE} \\
\cmidrule(lr){2-5}
\cmidrule(lr){6-9}
Region
& \betinn{} & TINN & PINN & PI-DeepONet
& \betinn{} & TINN & PINN & PI-DeepONet \\
\midrule

ATM (501)
& \textbf{.0345} & .0398 & .0390 & .0561
& \textbf{.0102} & .0124 & .0116 & .0174 \\

ATM tight (253)
& \textbf{.0328} & .0398 & .0384 & .0561
& \textbf{.0101} & .0123 & .0115 & .0171 \\

ATM wide (1297)
& \textbf{.0339} & .0392 & .0383 & .0590
& \textbf{.0100} & .0124 & .0115 & .0172 \\

ATM wide \& short maturity (62)
& .188 & \textbf{.181} & .182 & .307
& \textbf{.0131} & .0138 & .0133 & .0246 \\

ATM \& near maturity (8)
& .536 & .402 & \textbf{.376} & .476
& \textbf{.00849} & .0110 & .0114 & .0148 \\

\midrule

Long maturity (996)
& .244 & .266 & \textbf{.243} & .262
& \textbf{.00860} & .0100 & .00983 & .0108 \\

Near maturity (204)
& .0753 & .0806 & \textbf{.0751} & .214
& \textbf{.00201} & .00257 & .00231 & .00421 \\

Short maturity (520)
& \textbf{.0595} & .0704 & .0668 & .173
& \textbf{.00354} & .00413 & .00365 & .00730 \\

Very near maturity (109)
& .0928 & .0893 & \textbf{.0877} & .242
& \textbf{.00175} & .00218 & .00208 & .00370 \\

\midrule

Feller satisfied (8770)
& \textbf{.138} & .154 & .144 & .191
& \textbf{.00613} & .00733 & .00683 & .00956 \\

Feller violated (1230)
& \textbf{.174} & .182 & .183 & .258
& \textbf{.0142} & .0171 & .0164 & .0208 \\

High-$x$, high-$\tau$, high-$\theta$ (48)
& \textbf{.000527} & .000709 & .000639 & .000873
& .00944 & .00758 & \textbf{.00688} & .00862 \\

\midrule

Near $v_{\max}$ (554)
& \textbf{.150} & .151 & .154 & .190
& \textbf{.00517} & .00592 & .00551 & .00796 \\

Near $v_{\min}$ (520)
& \textbf{.171} & .187 & .175 & .247
& \textbf{.0117} & .0142 & .0131 & .0176 \\

Variance interior (4885)
& \textbf{.136} & .152 & .142 & .192
& \textbf{.00647} & .00775 & .00734 & .0100 \\

\midrule

$x>1.8$ (443)
& \textbf{.000744} & .000809 & .000785 & .000954
& \textbf{.0115} & .0119 & .0116 & .0163 \\

$x>1.9$ (224)
& \textbf{.000680} & .000739 & .000732 & .000812
& \textbf{.0117} & .0124 & .0120 & .0171 \\

\bottomrule
\end{tabular}%
}
\end{table*}

Tables~\ref{tab:regional_price_delta} and~\ref{tab:regional_kgamma_uv} show that relative model performance depends on both the local regime and the quantity being evaluated. \betinn{} is particularly consistent across the broad ATM subsets, where it attains the lowest mean error for all four primary quantities. The same pattern holds across the Feller-satisfied and Feller-violated partitions, the variance-interior subset, and neighborhoods of both variance boundaries.

The maturity-focused subsets expose a more metric-dependent picture. \betinn{} remains strongest for Price, $\Delta$, and $u_v$ across most of these regions, whereas PINN attains slightly lower $K\Gamma$ error in several near-maturity subsets. The small ATM--near-maturity intersection is particularly demanding for curvature, with substantially larger $K\Gamma$ errors than in the broader ATM regions.

The predefined high-$x$, high-$\tau$, high-$\theta$ stress subset provides a different failure mode: the best model depends on the observable, with TINN attaining the lowest Price error, PI-DeepONet the lowest $\Delta$ error, \betinn{} the lowest $K\Gamma$ error, and PINN the lowest $u_v$ error. Similarly, in the extreme high-moneyness subsets, Price and curvature errors favor \betinn{}, while the smallest $\Delta$ error can shift to PINN or TINN. These results show that no single architecture uniformly dominates every localized regime and observable.

Overall, the regional analysis reinforces the main-text conclusion that global error summaries do not fully characterize differential fidelity over the solution family. The spatial organization of these discrepancies is examined directly through the one- and two-dimensional diagnostics in Appendix~\ref{app:differential_diagnostics}.

\subsection{Additional TINN and PINN Baseline Analyses}
\label{app:additional_baselines}

The controlled final benchmark is complemented by two additional analyses addressing different aspects of baseline strength. We mainly examine whether moderate changes in the architecture of the TINN and PINN baselines materially alter their development performance. All candidates are trained and evaluated using the same six development seeds and the common controlled protocol described in Appendix~\ref{app:experimental_configuration}. 

\begin{table*}[t]
\centering
\caption{
Architecture sensitivity of the TINN and PINN baselines under approximately matched parameter budgets. All candidates are evaluated on the same six development seeds.
}
\label{tab:tuned_baselines}
\resizebox{\textwidth}{!}{%
\begin{tabular}{llrrrrrrrrl}
\toprule
Family
& Candidate
& Depth
& Widths
& Params
& Mismatch (\%)
& Price
& $\Delta$
& $K\Gamma$
& $u_v$
& Ranks / mean rank \\
\midrule

TINN
& Capacity-matched
& 2
& $(37,37)$
& 3663
& 1.89
& 0.007587
& 0.0179
& 0.148
& 0.00725
& 2/2/1/1 \quad 1.50 \\

TINN
& Funnel
& 2
& $(50,25)$
& 3585
& 0.28
& 0.007585
& 0.0175
& 0.150
& 0.00734
& 1/1/2/2 \quad 1.50 \\

TINN
& Depth 3
& 3
& $(26,26,26)$
& 3483
& 3.12
& 0.009363
& 0.0201
& 0.158
& 0.00873
& 3/3/3/3 \quad 3.00 \\

\midrule

PINN
& Capacity-matched
& 2
& $(55,55)$
& 3631
& 1.00
& 0.00859
& 0.01868
& 0.1489
& 0.00816
& 1/2/3/1 \quad 1.75 \\

PINN
& Depth 3
& 3
& $(39,39,39)$
& 3511
& 2.34
& 0.00997
& 0.02016
& 0.1448
& 0.00900
& 3/3/1/3 \quad 2.50 \\

PINN
& Depth 4
& 4
& $(33,33,33,33)$
& 3697
& 2.84
& 0.00893
& 0.01866
& 0.1451
& 0.00821
& 2/1/2/2 \quad 1.75 \\

\bottomrule
\end{tabular}%
}
\end{table*}

Table~\ref{tab:tuned_baselines} shows that moderate changes in baseline geometry do not produce a uniformly stronger alternative across the four primary quantities. For TINN, the capacity-matched and funnel configurations trade small improvements between Price and $\Delta$ versus $K\Gamma$ and $u_v$, while the deeper configuration is consistently weaker. For PINN, additional depth can improve selected derivative metrics, particularly $K\Gamma$, but does not improve the joint error profile across Price, $\Delta$, $K\Gamma$, and $u_v$.

These results reduce concern that the main benchmark depends on an under-capacity or poorly chosen standard baseline architecture. The lightweight search reveals metric-level trade-offs within each baseline family without producing a configuration that uniformly dominates the capacity-controlled baseline used in the final comparison.

\subsection{DeepSVM as an External Heston-Specific Reference}
\label{app:deepsvm}

\paragraph{DeepSVM-adapted.}
DeepSVM \citep{malandain2025deepsvm} is a physics-informed DeepONet developed specifically for the Heston pricing operator under a specific training and evaluation protocol. We implement an independent DeepSVM adapted reference using branch inputs $(\kappa,\theta,\sigma,\rho,r)$ and trunk inputs $(x,v,\tau)$, four residual GELU layers, and the published hard terminal construction
\[
u_{\mathrm{pred}}
=
\max(e^x-1,0)
+
\tau\,
\mathrm{Softplus}
\!\left(
\langle b(\kappa,\theta,\sigma,\rho,r),
        t(x,v,\tau)\rangle
\right).
\]
The branch and trunk widths and latent dimension are set to $64$, yielding $33{,}920$ trainable parameters. Training retains the published $R^2+0.1R^4$ physics objective, dedicated ATM residual sampling, residual-based adaptive refinement, and Adam--L-BFGS optimization. We use $200{,}000$ active collocation points, $4{,}096$ ATM points, $2{,}048$ boundary points during Adam, an additional $2{,}048$ boundary points before L-BFGS, and residual-based refinement every $500$ Adam steps using $50{,}000$ candidates and $20{,}000$ replacements. Optimization uses $10{,}000$ Adam iterations with initial learning rate $10^{-4}$ halved every $2{,}000$ steps, followed by up to $5{,}000$ L-BFGS iterations with memory size $20$.

To evaluate DeepSVM on the same parametric Heston problem considered throughout this work, its sampling ranges are extended to cover the corresponding evaluation distribution using
\[
\tau\in[0,5],\qquad
x\in[-2,2],\qquad
v\in[0.1,1],
\]
\[
r\in[0,0.2],\qquad
\rho\in[-0.9,0.1],\qquad
\kappa\in[0.5,3],\qquad
\theta\in[0.01,1],\qquad
\sigma\in[0.1,1].
\]
Input normalization uses these same bounds, and the training distribution contains both Feller-satisfied and Feller-violated configurations. Checkpoint selection uses validation physics loss, and the selected checkpoint is reloaded from disk before evaluation. All reported metrics use the same frozen $10{,}000$-point final panel and the same semi-analytic Heston oracle used for the controlled benchmark, with Greeks obtained by automatic differentiation of the trained network.

\begin{table}[t]
\centering
\caption{External adapted DeepSVM reference. All models are evaluated on the same frozen $10{,}000$-point panel against the same semi-analytic oracle. Values are mean $\pm$ std over seeds.
}
\label{tab:deepsvm_adapted}
\resizebox{\linewidth}{!}{%
\begin{tabular}{lrrcccc}
\toprule
Method & Params & Price MAE & $\Delta$ MAE & $K\Gamma$ MAE & $u_v$ MAE \\
\midrule
DeepSVM-adapted
& 33920 
& 0.1258 $\pm$ 0.0004
& 0.2073 $\pm$ 0.0003
& 0.376 $\pm$ 0.002
& 0.0567 $\pm$ 0.0004 \\
\midrule
\betinn{}
& 3595 
& 0.00746 $\pm$ 0.00097
& 0.0180 $\pm$ 0.0010
& 0.143 $\pm$ 0.005
& 0.00712 $\pm$ 0.00089 \\
TINN
& 3663 
& 0.00910 $\pm$ 0.00062
& 0.0197 $\pm$ 0.0009
& 0.158 $\pm$ 0.006
& 0.00853 $\pm$ 0.00046 \\
PINN
& 3631 
& 0.00857 $\pm$ 0.00054
& 0.0188 $\pm$ 0.0005
& 0.149 $\pm$ 0.003
& 0.00800 $\pm$ 0.00049 \\
PI-DeepONet
& 3511 
& 0.00974 $\pm$ 0.00063
& 0.0232 $\pm$ 0.0007
& 0.199 $\pm$ 0.016
& 0.0109 $\pm$ 0.0005 \\
\bottomrule
\end{tabular}%
}
\end{table}

\paragraph{Comparison and derivative-level behavior.}
Table~\ref{tab:deepsvm_adapted} places DeepSVM-adapted alongside the controlled benchmark as an external Heston-specific reference. The controlled architectures achieve substantially smaller errors on the common evaluation panel despite using a smaller number of parameters. The difference is largest for the normalized option value and Delta, while the gap is narrower for $K\Gamma$. 

The regional diagnostics localize the main weakness of DeepSVM-adapted around the at-the-money region, where both pricing and derivative errors are most pronounced. The degradation becomes stronger as the analysis focuses more tightly on ATM, while performance is markedly more stable near maturity. This concentration of error is consistent with the sensitivity artifacts reported in the original DeepSVM study and indicates that the model's main difficulty is not uniform across the domain, but localized in the region where the payoff geometry and its derivatives are most challenging.

The hard terminal construction provides one structural source of difficulty for derivative accuracy near the money. Because the fixed payoff term $\max(e^x-1,0)$ retains its kink at $x=0$, while the learned correction is smooth in $x$, the resulting representation inherits the payoff derivative discontinuity. This is consistent with the elevated ATM $\Delta$ error observed in our adapted implementation. For $K\Gamma$ the relationship is less direct, the payoff contribution to $u_{xx}-u_x$ vanishes pointwise away from the kink, so the observed ATM curvature error also reflects the accuracy and smoothness of the learned correction. This interpretation is consistent with the derivative-level instability reported for the original DeepSVM formulation \citep{malandain2025deepsvm}.

Our optimization diagnostics indicate that these errors persist after substantial reduction of the DeepSVM objective. Across seeds, L-BFGS executes the full $5{,}000$-iteration budget and reduces the post-Adam objective by approximately $93$--$95\%$. Validation physics loss also improves strongly after the second optimization stage. The remaining predictive and derivative errors therefore characterize the trained DeepSVM-adapted solution under its specified objective and structural constraints, and are not explained by the premature L-BFGS termination observed in our initial implementation.

\section{Local and Spatial Differential Diagnostics}
\label{app:differential_diagnostics}

This section complements the global and regional metrics in Appendix~\ref{app:final_benchmark}  with one-dimensional derivative profiles, two-dimensional error maps, and spatial PDE-residual diagnostics across representative interior, near-maturity, and stress regimes.

\subsection{Local Derivative Profiles}
\label{app:local_derivative_diagnostics}

\begin{figure*}[t]
    \centering
    \includegraphics[width=0.9\textwidth]
    {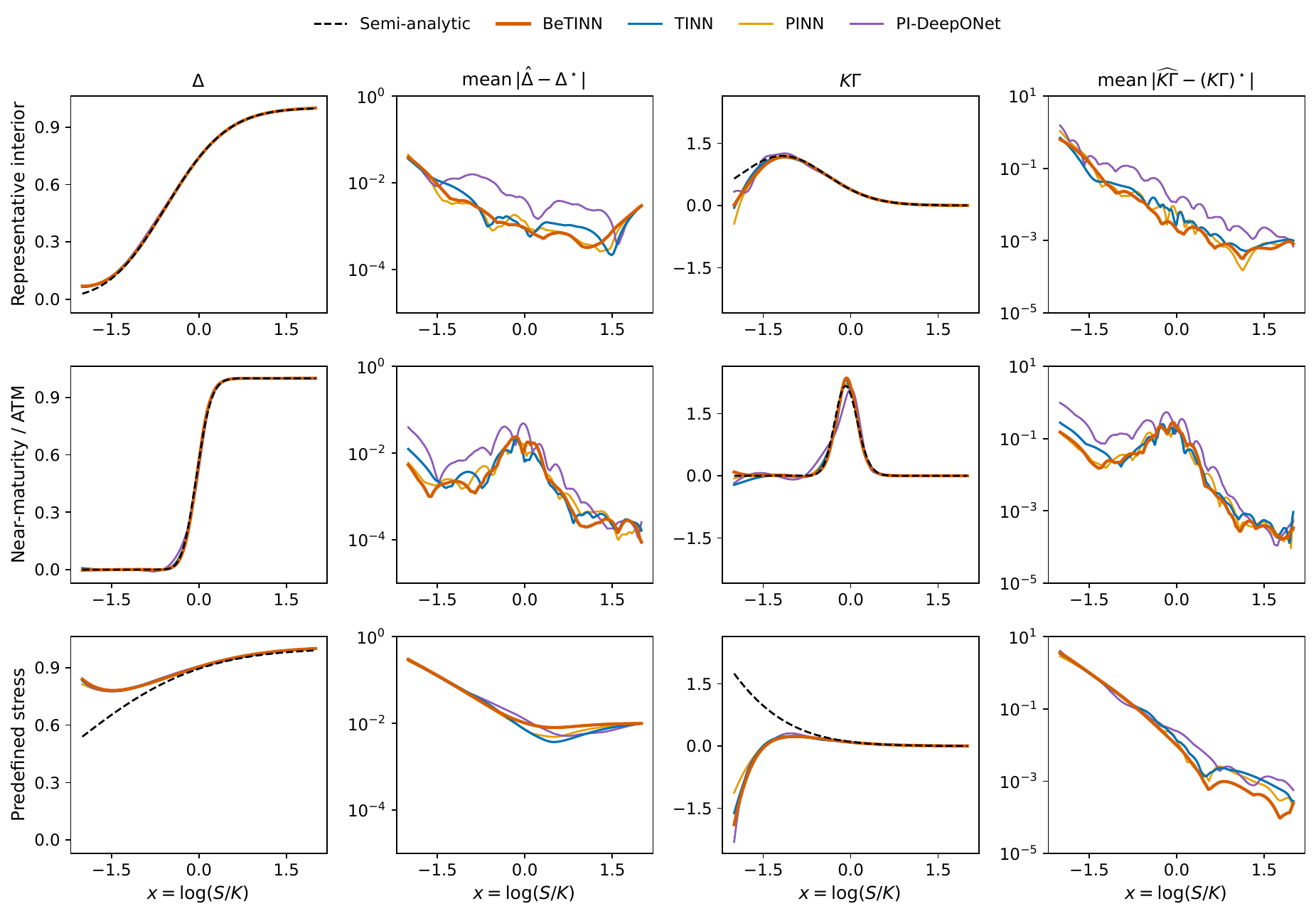}
    \caption{
    \textbf{Local Delta and Gamma profiles and pointwise errors.}
    Delta $\Delta$, absolute Delta error, strike-scaled Gamma $K\Gamma$, and absolute Gamma error are shown as functions of log-moneyness across representative interior, near-maturity, and stress regimes. Error panels use logarithmic scales to expose localized discrepancies spanning several orders of magnitude.
    }
    \label{fig:derivative_profile_errors}
\end{figure*}

We first examine one-dimensional slices through the same representative interior, near-maturity, and stress regimes used in the main-text qualitative analysis. Figure~\ref{fig:derivative_profile_errors} shows that derivative fidelity is regime and location dependent. In the interior regime, all models recover the principal shapes of both $\Delta$ and $K\Gamma$, although the error panels reveal localized differences that are difficult to infer from the profiles alone. Near maturity, errors concentrate around regions of sharp curvature, particularly around the localized $K\Gamma$ peak. In the stress regime, discrepancies in both first- and second-order sensitivities remain visible even where the corresponding option-value profiles are comparatively close.

The large low-moneyness errors visible in $\Delta$ and $K\Gamma$ do not correspond to larger derivative errors in the normalized coordinate system. On the frozen confirmatory panel, the errors in $u_x$ and $u_{xx}-u_x$ within $x<-1.5$ and $x<-1.8$ are no larger than their global errors for any of the four benchmark architectures. The growth in the reported financial sensitivities is therefore explained by the coordinate factors $e^{-x}$ and $e^{-2x}$ entering $\Delta$ and $K\Gamma$, respectively.

\begin{figure*}[t]
    \centering
    \includegraphics[width=0.8\textwidth]{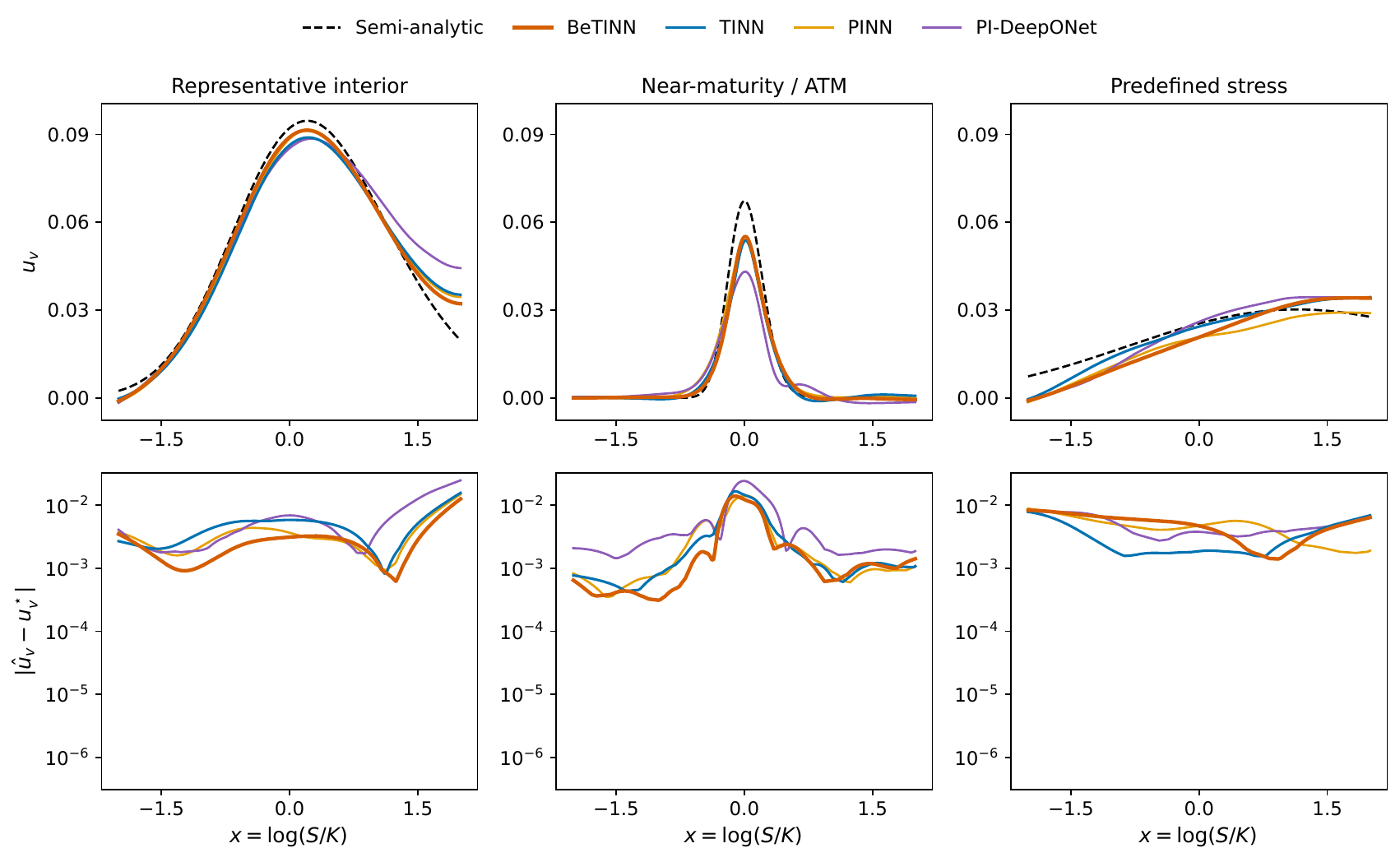}
    \caption{
    \textbf{Variance-sensitivity profiles across representative regimes.}
    The variance sensitivity and its absolute error are shown as functions of log-moneyness for the representative interior, near-maturity, and stress regimes. The error panels use a logarithmic scale to resolve differences across several orders of magnitude.
    }
    \label{fig:uv_profiles}
\end{figure*}

The same analysis is extended to variance sensitivity in
Figure~\ref{fig:uv_profiles}. Because $u_v$ differentiates the learned surface
along the variance coordinate, it probes a direction of the solution geometry
that is distinct from the spot-direction derivatives $\Delta$ and $K\Gamma$.

As shown in Figure~\ref{fig:uv_profiles}, the models reproduce the principal qualitative structure of the reference $u_v$ profiles, while the magnitude and location of the errors vary across regimes. Near maturity, discrepancies are concentrated around the localized sensitivity structure, whereas the stress regime exhibits a broader error profile across log-moneyness. These results complement the global $u_v$ metrics in Table~\ref{tab:global_metrics} by showing how aggregate error is distributed along representative solution slices.

\subsection{Spatial Structure of Derivative Errors}
\label{app:spatial_derivative_errors}

To characterize the joint dependence of derivative error on both log-moneyness and variance, we evaluate $\Delta$, $K\Gamma$, and $u_v$ over common dense $(x,v)$ grids. Within each diagnostic, all methods and regimes are evaluated on the same $55\times55$ grid and displayed using a shared logarithmic color scale.

\begin{figure*}[t]
    \centering
    \includegraphics[width=0.7\textwidth]{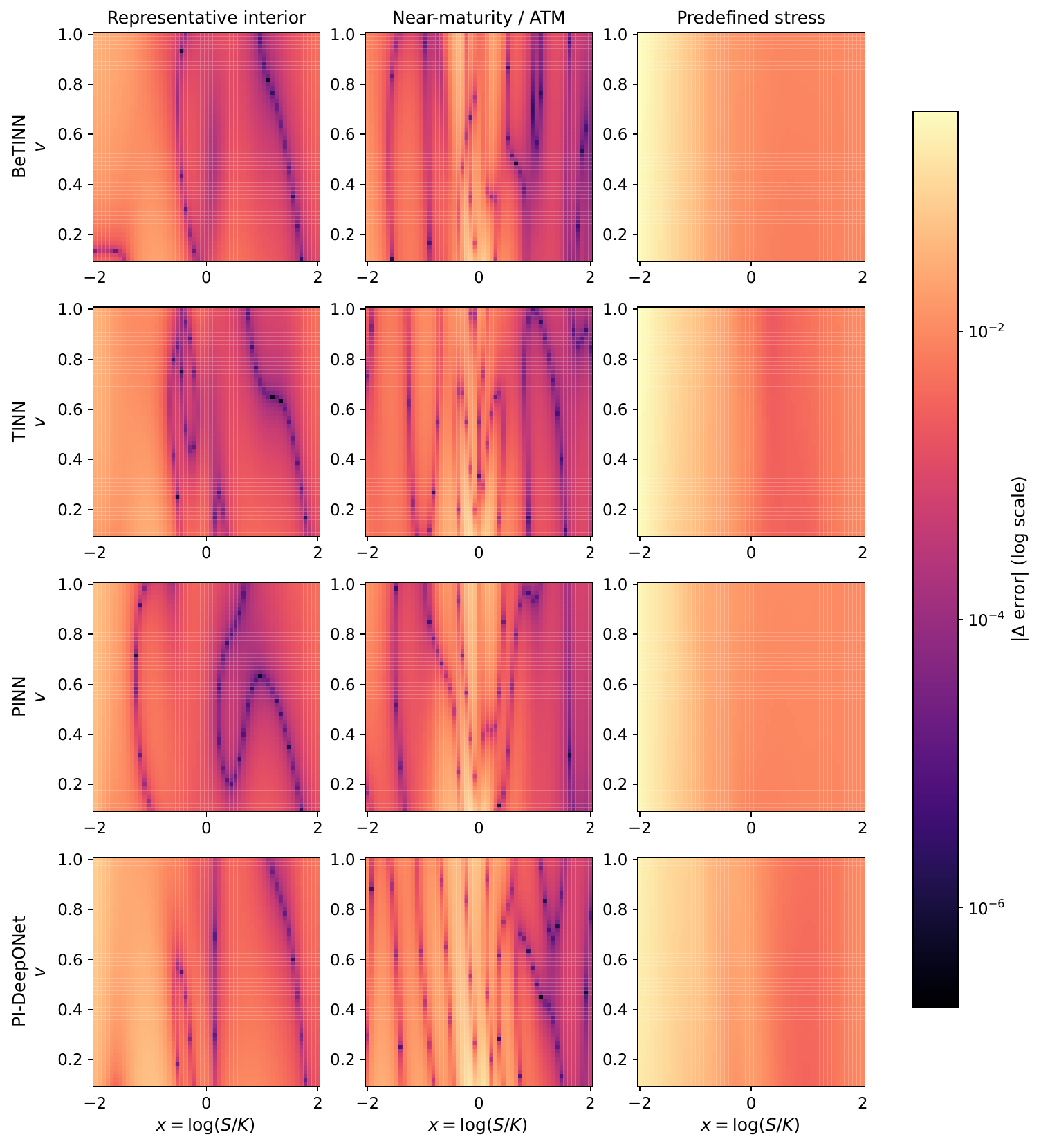}
    \caption{
    \textbf{Spatial distribution of Delta error.}
    Absolute pointwise Delta error,
    $|\hat{\Delta}-\Delta^\star|$, is shown over the $(x,v)$ plane across representative interior, near-maturity, and stress regimes. All methods are evaluated on the same
    $55\times55$ grid, and all panels share a common logarithmic color scale. Narrow low-error structures correspond to locations of small pointwise error.
    }
    \label{fig:delta_error_maps}
\end{figure*}

Figure~\ref{fig:delta_error_maps} demonstrates that Delta error is not spatially uniform. The geometry of the error field changes across both regimes and architectures, indicating that differences between methods cannot be described solely as a uniform rescaling of global Delta MAE.

\begin{figure*}[t]
    \centering
    \includegraphics[width=0.7\textwidth]{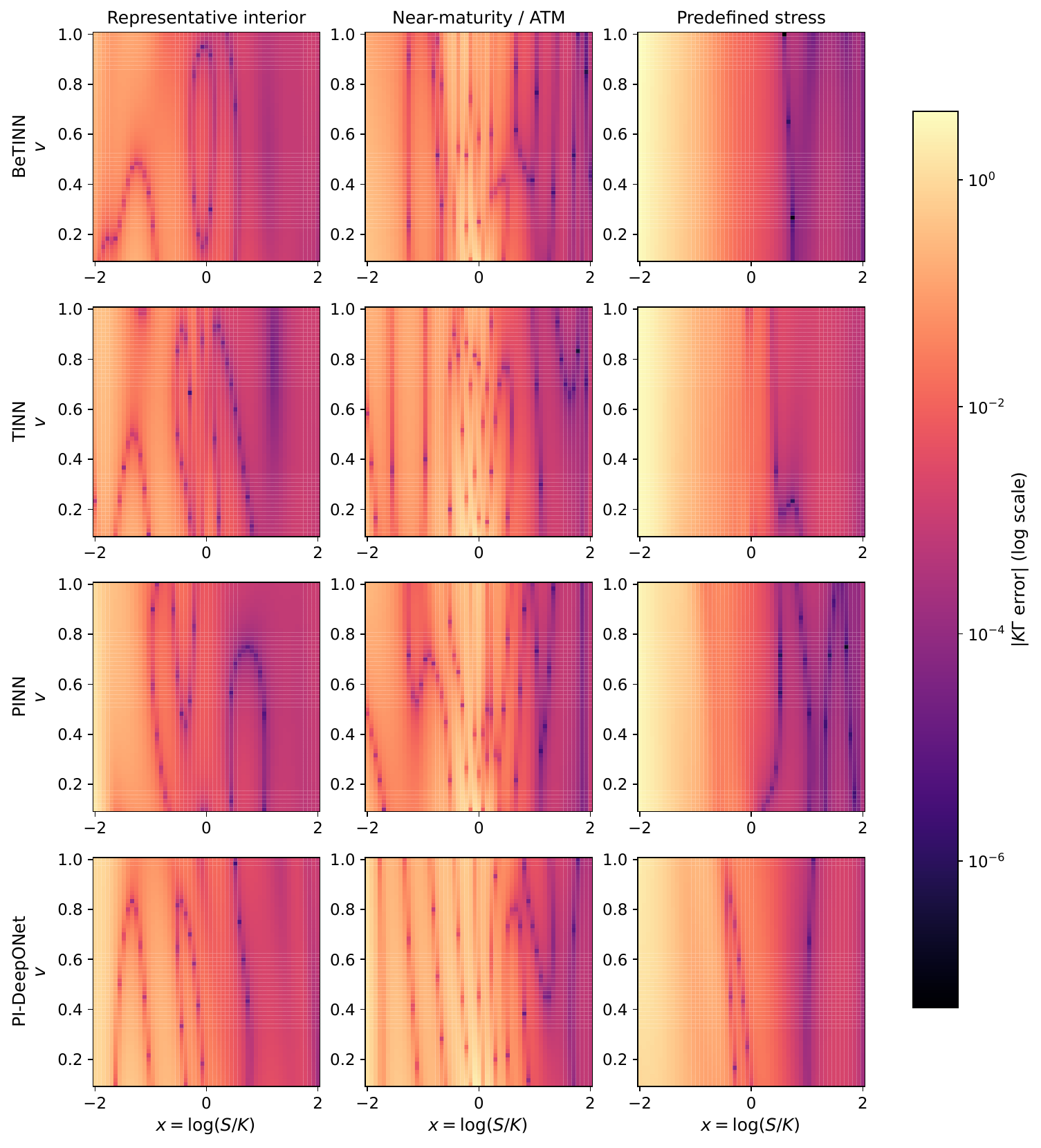}
    \caption{
    \textbf{Spatial distribution of strike-scaled Gamma error.}
    Absolute pointwise error in $K\Gamma$ is shown over the $(x,v)$ plane across representative interior, near-maturity,
    and stress regimes. All panels use the same $55\times55$ evaluation grid
    and a shared logarithmic color scale. Narrow low-error structures
    correspond to locations where the pointwise error approaches zero.
    }
    \label{fig:kgamma_error_maps}
\end{figure*}

Second-order sensitivity provides a stricter local diagnostic because $K\Gamma$ depends on the curvature of the learned solution. As shown in Figure~\ref{fig:kgamma_error_maps}, $K\Gamma$ error exhibits pronounced spatial heterogeneity, particularly in the near-maturity regime. This behavior is consistent with the local profiles in Figure~\ref{fig:derivative_profile_errors}, where sharp curvature produces localized second-order discrepancies that are not apparent from option-value accuracy alone.

\begin{figure*}[t]
    \centering
    \includegraphics[width=0.7\textwidth]{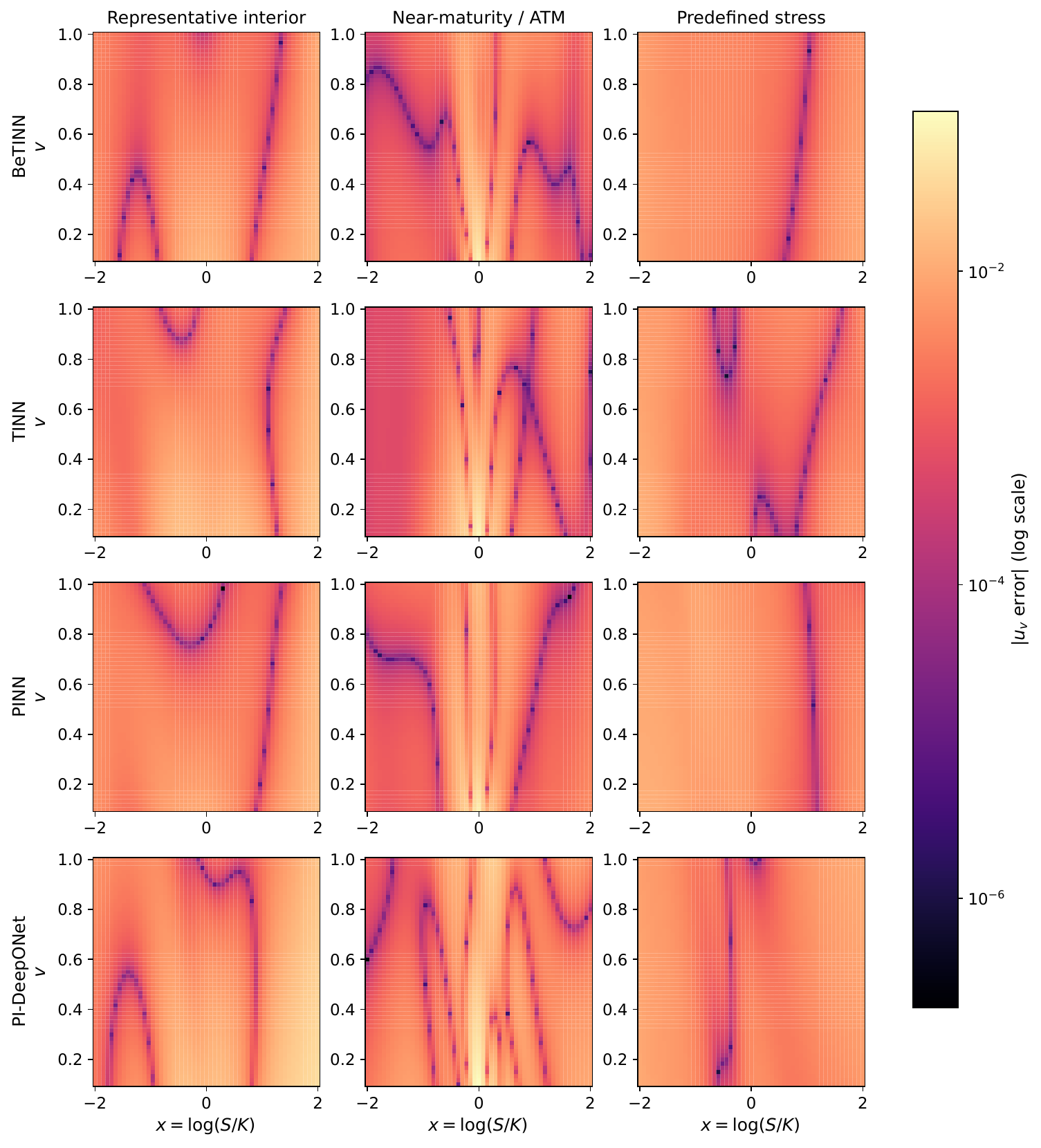}
    \caption{
    \textbf{Spatial distribution of variance-sensitivity error.}
    Absolute pointwise variance-sensitivity error,
    $|\hat{u}_v-u_v^\star|$, is shown over the $(x,v)$ plane across representative interior, near-maturity,
    and stress regimes. All panels use the same $55\times55$ evaluation grid
    and a shared logarithmic color scale.
    }
    \label{fig:uv_error_maps}
\end{figure*}

Figure~\ref{fig:uv_error_maps} examines the corresponding spatial structure of variance-sensitivity error. It confirms that $u_v$ error depends jointly on log-moneyness and variance rather than only on the one-dimensional slices in Figure~\ref{fig:uv_profiles}. Taken together, Figures~\ref{fig:delta_error_maps}--\ref{fig:uv_error_maps} show that global derivative metrics summarize error surfaces with substantial spatial structure, motivating the use of both aggregate and local diagnostics when assessing differential fidelity.

\subsection{Spatial Structure of the PDE Residual}
\label{app:pde_residual_structure}

\begin{figure*}[t]
    \centering
    \includegraphics[width=0.7\textwidth]{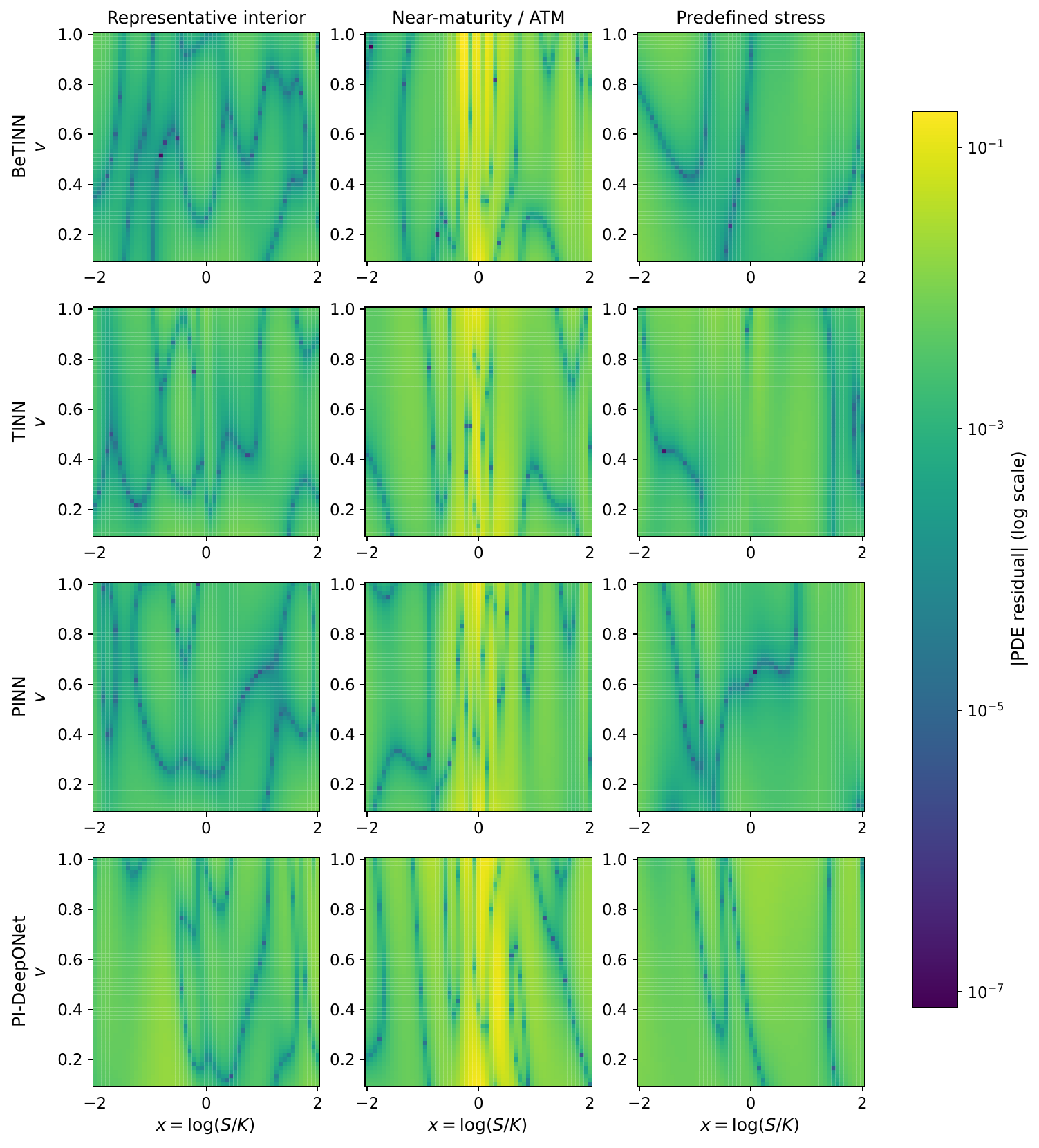}
    \caption{
    \textbf{Spatial distribution of the Heston PDE residual.}
    Absolute pointwise PDE residuals are shown over the $(x,v)$ plane across representative interior, near-maturity,
    and stress regimes. Residuals are evaluated on the same $55\times55$
    grids used for the spatial derivative diagnostics and displayed using a
    common logarithmic color scale.
    }
    \label{fig:pde_residual_maps}
\end{figure*}

Derivative error and PDE-residual error quantify different properties of a learned surrogate. The former measures agreement with the semi-analytic reference derivatives, whereas the latter measures local consistency with the governing differential equation. To compare these criteria directly, Figure~\ref{fig:pde_residual_maps} evaluates the absolute PDE residual over the same class of dense $(x,v)$ grids.

Figure~\ref{fig:pde_residual_maps} shows structured residual fields across methods and regimes. The near-maturity regime exhibits particularly localized variation, while the interior and stress regimes display distinct method-dependent spatial patterns. Comparison with Figures~\ref{fig:delta_error_maps}--\ref{fig:uv_error_maps} further shows that small PDE residual and small derivative error are not pointwise identical criteria. This spatial distinction is consistent with the run-level relationship between validation physics loss and downstream error reported in Table~\ref{tab:validation_spearman}.

\subsection{Extended Sensitivity Profiles}
\label{app:extended_sensitivity_results}

\begin{figure*}[t]
    \centering
    \includegraphics[width=0.8\textwidth]
    {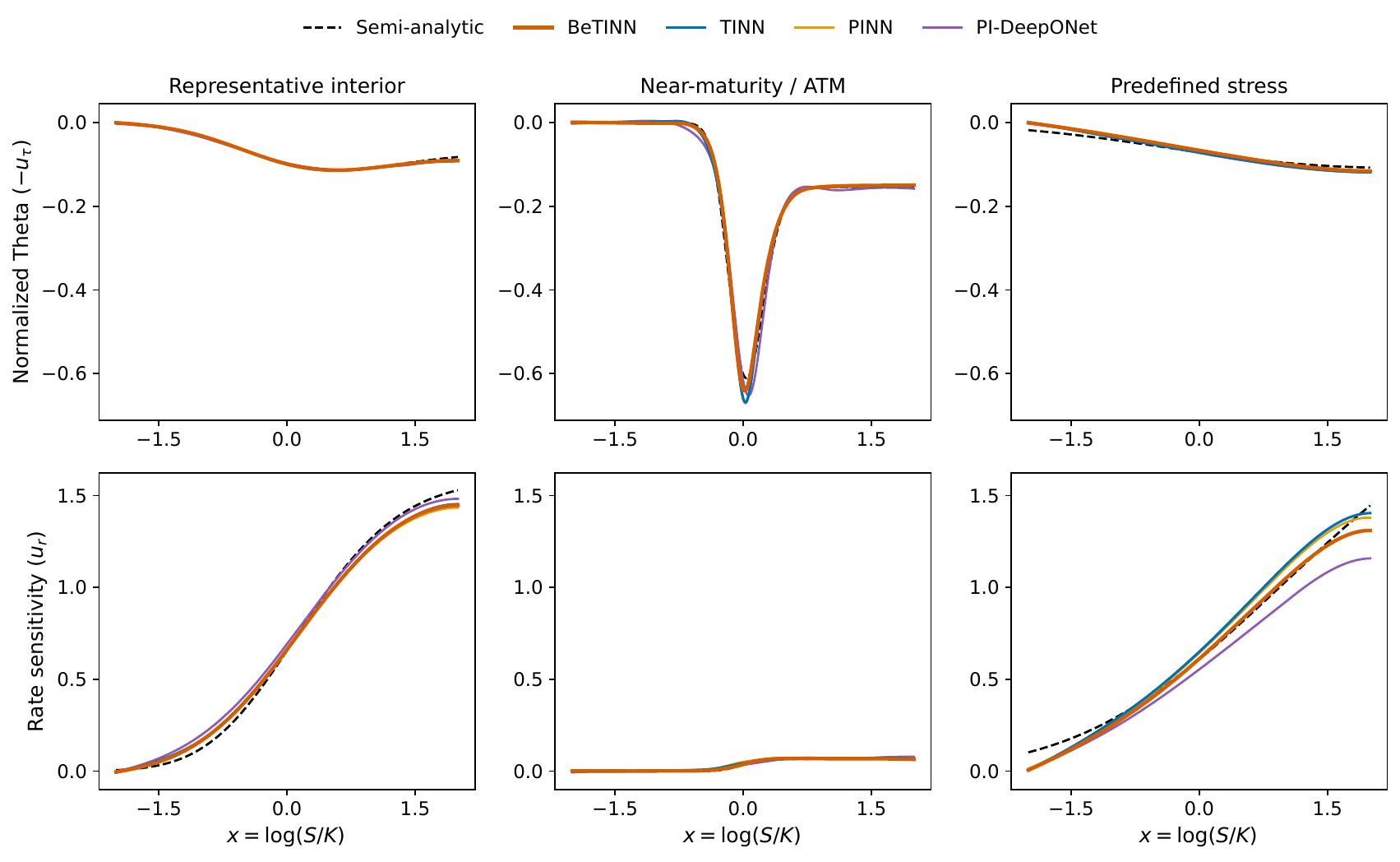}
    \caption{
    \textbf{Extended sensitivity profiles across representative regimes.}
    Normalized calendar Theta $-u_\tau$ and interest-rate sensitivity $u_r$
    are shown as functions of log-moneyness $x=\log(S/K)$ for representative
    interior, near-maturity, and stress regimes.
    }
    \label{fig:extended_sensitivities}
\end{figure*}

The preceding diagnostics focus on $\Delta$, $K\Gamma$, and $u_v$. Figure~\ref{fig:extended_sensitivities} extends the local analysis to normalized calendar Theta $-u_\tau$ and interest-rate sensitivity $u_r$. Both quantities are obtained by automatic differentiation of the same trained surrogates, without separately trained derivative models.

Figure~\ref{fig:extended_sensitivities} shows close qualitative agreement with the semi-analytic reference across most interior and near-maturity profiles. The stress regime is more discriminating, particularly for $u_r$, where differences between learned surrogates and the reference become more visible. Together with the aggregate results in Table~\ref{tab:global_metrics}, these profiles show that the differential evaluation extends beyond the primary spot- and variance-direction sensitivities.

\section{Shape and Sign-Constraint Diagnostics}
\label{app:shape_diagnostics}

Low predictive error does not by itself guarantee that a learned option surface preserves economically relevant local shape properties. We therefore complement the error-based evaluation with violation rates for monotonicity, convexity, variance sensitivity, calendar-time sensitivity, and interest-rate sensitivity. These diagnostics are evaluated on the same frozen final panel used for the confirmatory benchmark.

\begin{figure}[t]
    \centering
    \includegraphics[width=0.8\linewidth]{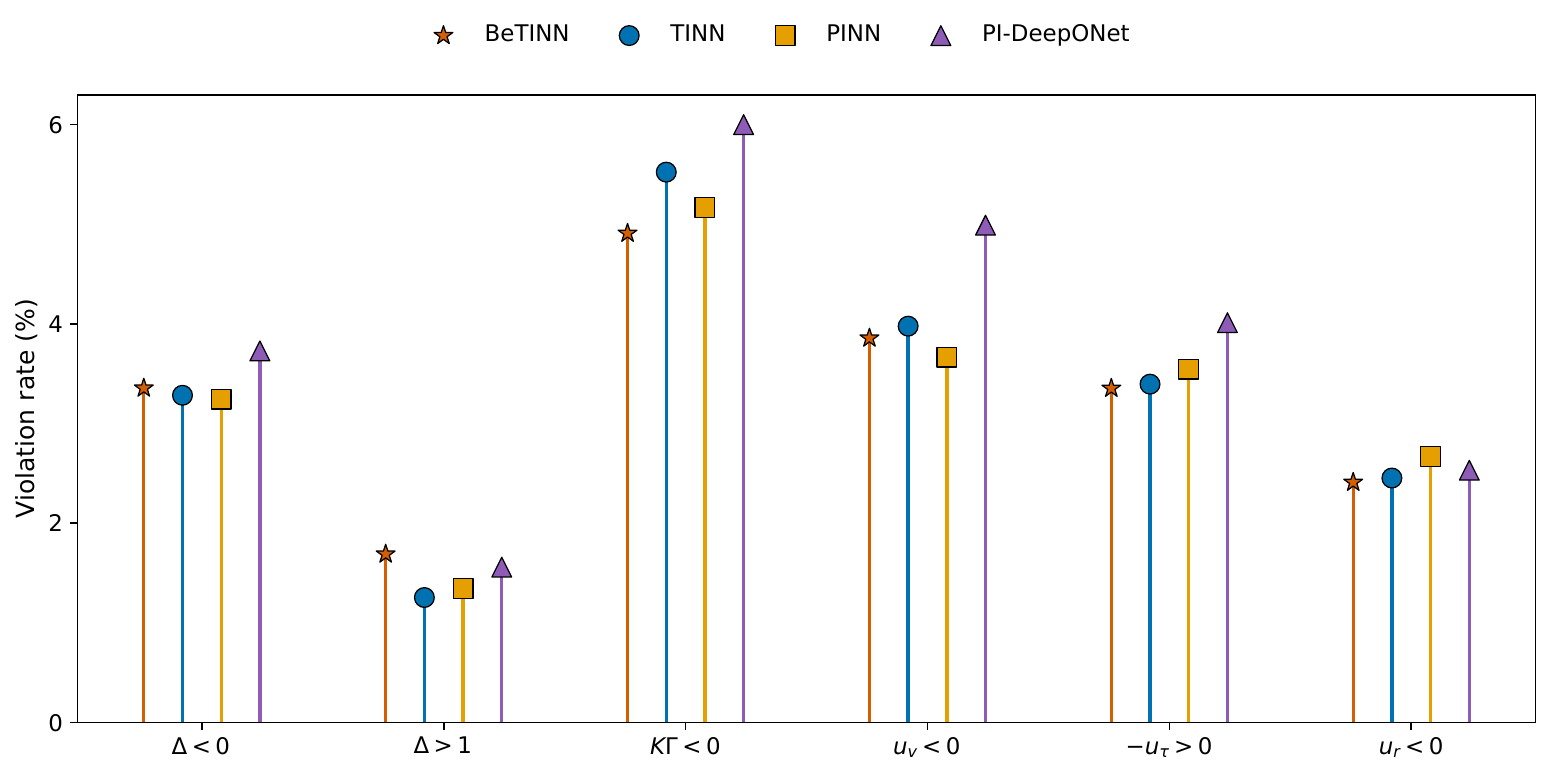}
    \caption{
\textbf{Shape-constraint violation rates.}
Percentage of points on the frozen confirmatory panel violating the primary
structural conditions using tolerance $\varepsilon=10^{-4}$. The corresponding
semi-analytic oracle has zero violations under the same tolerance. Lower values
indicate fewer violations.
}
    \label{fig:shape_diagnostics}
\end{figure}

Figure~\ref{fig:shape_diagnostics} shows that all learned architectures retain nonzero violation rates after applying a tolerance of $\varepsilon=10^{-4}$, while the semi-analytic oracle has zero violations under the same criterion. The rates are therefore not explained by numerical ambiguity near the constraint boundaries. The relative ordering remains condition dependent, and improved aggregate predictive accuracy therefore does not imply uniform preservation of monotonicity, convexity, or sensitivity signs.

Importantly, these properties are evaluated rather than explicitly enforced by the training objective. The reported violation rates therefore characterize the structural behavior of the learned surrogates but do not constitute a formal guarantee of arbitrage consistency.

\section{Cross-PDE Transfer Checks}
\label{app:cross_pde}

We evaluate \betinn{} with the configuration used for Heston on two additional parametric PDE families. We consider parametric extensions of the Burgers and Wave equations used in  \citet{dai2026tinns} to compare PINN, TINN, and \betinn{} under a common training and evaluation protocol. No PDE-specific architectural tuning is applied.

\paragraph{Parametric PDEs.}
For Burgers, we vary the viscosity through
\begin{equation}
    u_t + u u_x - \nu(\lambda) u_{xx} = 0,
    \qquad
    \nu(\lambda)=\lambda\frac{0.01}{\pi},
    \qquad
    \lambda\in[0.5,2],
\end{equation}
on $x\in[-1,1]$ and $t\in[0,1]$, with the original initial and boundary conditions. The benchmark used in  \citet{dai2026tinns} is recovered at $\lambda=1$.

For Wave, we consider
\begin{equation}
    u_{tt}-c^2u_{xx}=0,
    \qquad
    c\in[1,3],
\end{equation}
on $x,t\in[0,1]$, with the same initial and boundary conditions as in  \citet{dai2026tinns}. Its exact parametric solution is
\begin{equation}
    u(x,t;c)
    =
    \sin(\pi x)\cos(c\pi t)
    +
    \frac{1}{2}\sin(4\pi x)\cos(4c\pi t),
\end{equation}
and the original Wave problem corresponds to $c=2$.

All models are trained in 64-bit precision with the same Levenberg--Marquardt protocol and checkpoint selection rule. Training panels are shared across architectures within each seed. Final evaluation uses frozen dense $(x,t,\mu)$ panels and six seeds. 

\begin{table}[t]
\centering
\caption{
Errors over the additional frozen parametric PDE evaluation panels, reported as mean $\pm$ std over six seeds.
}
\label{tab:cross_pde_transfer}
\begin{tabular}{llrcc}
\toprule
PDE & Model & Params & Rel.\ $L_2$ & MAE \\
\midrule
Parametric Burgers
& PINN   & 3017 & $0.04811 \pm 0.0560$ & $0.003818 \pm 0.0062$ \\
& TINN   & 2985 & $0.003053 \pm 0.0035$ & $0.000538 \pm 0.0011$ \\
& \betinn{} & 2967 & $0.000379 \pm 0.00023$ & $0.00000825 \pm 0.0000037$ \\
\midrule
Parametric Wave
& PINN   & 3017 & $0.002724 \pm 0.0035$ & $0.0000453 \pm 0.000049$ \\
& TINN   & 2985 & $0.002101 \pm 0.0021$ & $0.0000717 \pm 0.000076$ \\
& \betinn{} & 2967 & $0.004134 \pm 0.0056$ & $0.0000733 \pm 0.000089$ \\
\bottomrule
\end{tabular}
\end{table}

\paragraph{Cross-PDE performance.}
Table~\ref{tab:cross_pde_transfer} shows substantially lower mean errors for \betinn{} than for both baselines on parametric Burgers, showing that the conditional architecture remains effective outside the Heston setting for a nonlinear dissipative system. The Wave family is less favorable to the same fixed configuration and the ordering varies by metric. \betinn{} does not retain the advantage observed on Heston and Burgers.

These results indicate that the effectiveness of \betinn{} depends on the structure of the parametric solution family. The Burgers results are consistent with richer conditional adaptation being useful when the solution varies substantially with the physical parameter. The Wave results identify a regime in which the fixed \(q=4\) configuration provides no systematic advantage over simpler alternatives.

\paragraph{Wave parameter diagnostic.}
We examine the Wave error across all $21$ values of $c\in\{1.0,1.1,\ldots,3.0\}$ in the frozen evaluation panel. For all three models error increases sharply near the upper end of the parameter interval and the maximum relative $L_2$ error and MAE occur at $c=3$. The interval $c\geq2.6$ accounts for $97$--$100\%$ of the squared prediction error in each run. The nominal case $c=2$ remains substantially easier; for \betinn{}, its mean relative $L_2$ error at $c=2$ is approximately $3.0\times10^{-5}$, compared with a global mean of $4.1\times10^{-3}$.

This concentration is consistent with the increasing temporal frequency of the exact solution,
\begin{equation}
    \cos(c\pi t)
    \quad\text{and}\quad
    \cos(4c\pi t),
\end{equation}
as $c$ approaches the upper end of the interval. 

We additionally ran a diagnostic on Wave to assess sensitivity to \betinn{}'s modulation dimensionality. Reducing $q$ from $4$ to $2$ decreased the mean relative $L_2$ error to $0.00402$ and the mean MAE to $0.0000574$. The lower-$q$ configuration also reduced the error in the high-$c$ regime and used fewer parameters. These results indicate that specific problems and applications tuning of conditional capacity can yield additional gains. We retain $q=4$ in Table~\ref{tab:cross_pde_transfer} to preserve the same \betinn{} specification across Heston, Burgers, and Wave. 

Overall, the cross-PDE experiments show strong performance on the parametric Burgers family, while Wave exposes greater sensitivity to the choice of conditional capacity under the fixed, untuned configuration. Heston remains our primary benchmark as it combines a heterogeneous parameter space, mixed derivatives, and downstream quantities whose accuracy depends directly on first and second order solution derivatives. This setting provides a demanding test of the differential structure learned by a physics-informed surrogate. Our results indicate that \betinn{} is useful for parametric PDEs with strong time--parameter variation and demanding sensitivity requirements. In applications with these conditions, its richer conditional adaptation provides an effective alternative to standard PINNs and temporally conditioned architectures.

\end{document}